\documentclass[10pt,twocolumn,letterpaper]{article}
\usepackage{iccv}              
\usepackage{tabularx}
\usepackage{pifont}
\usepackage{adjustbox}
\usepackage{multirow}
\usepackage{booktabs}
\usepackage{graphicx}
\usepackage{colortbl}
\title{GenCLIP: Generalizing CLIP Prompts for Zero-shot Anomaly Detection}

\author{
Donghyeong Kim\thanks{Equal contribution.} \quad 
Chaewon Park\footnotemark[1] \quad 
Suhwan Cho\quad 
Hyeonjeong Lim \\
Minseok Kang \quad 
Jungho Lee\quad 
Sangyoun Lee \\
Yonsei University, Seoul, Korea \\
{\tt\small \{2donghyung87, chaewon28, chosuhwan, jeonggg119, louis0503, 2015142131, syleee\}@yonsei.ac.kr}
}

\begin{document}
\maketitle
\begin{abstract} 

Zero-shot anomaly detection (ZSAD) aims to identify anomalies in unseen categories by leveraging CLIP's zero-shot capabilities to match text prompts with visual features. A key challenge in ZSAD is learning general prompts stably and utilizing them effectively, while maintaining both generalizability and category specificity. Although general prompts have been explored in prior works, achieving their stable optimization and effective deployment remains a significant challenge. In this work, we propose GenCLIP, a novel framework that learns and leverages general prompts more effectively through multi-layer prompting and dual-branch inference. Multi-layer prompting integrates category-specific visual cues from different CLIP layers, enriching general prompts with more comprehensive and robust feature representations. By combining general prompts with multi-layer visual features, our method further enhances its generalization capability. To balance specificity and generalization, we introduce a dual-branch inference strategy, where a vision-enhanced branch captures fine-grained category-specific features, while a query-only branch prioritizes generalization. The complementary outputs from both branches improve the stability and reliability of anomaly detection across unseen categories. Additionally, we propose an adaptive text prompt filtering mechanism, which removes irrelevant or atypical class names not encountered during CLIP's training, ensuring that only meaningful textual inputs contribute to the final vision-language alignment. 

\vspace{-2mm}

\end{abstract}    
\section{Introduction}
\begin{figure}[t]
\begin{center}
    \includegraphics[width=1\linewidth]{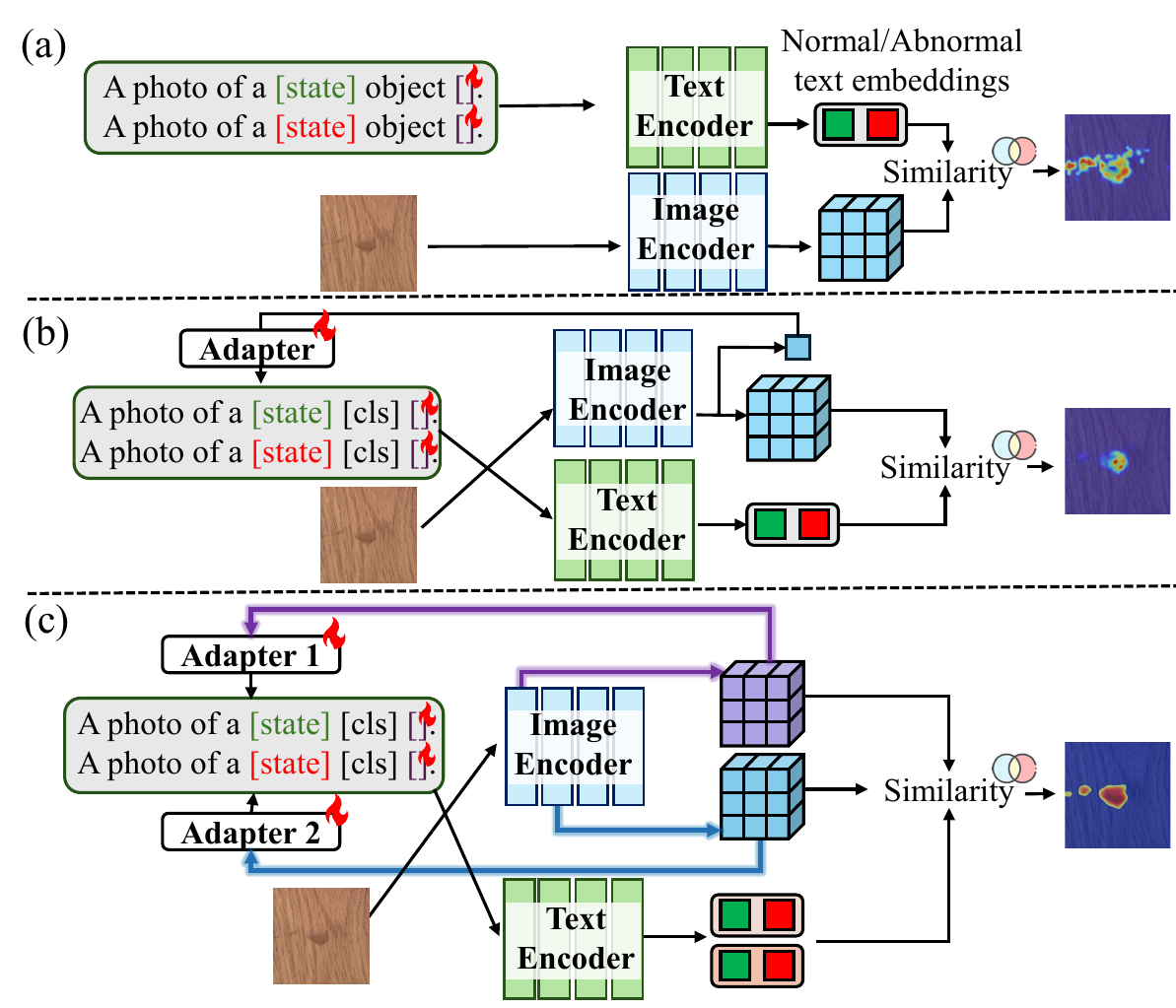}
\end{center}    
\vspace{-4mm}
   \caption{Paradigms of CLIP prompt learning-based zero-shot anomaly detection. (a) Methods that make generalizable query text prompts (b) Methods that adapt high-level CLIP vision features to facilitate text prompts. (c) Our approach that leverages multi-layer features from CLIP visual encoder to augment text embeddings.}
\vspace{-4mm}
\label{fig:fig1}
\vspace{-2mm}

\end{figure}

Anomaly detection (AD) is a critical research field that focuses on identifying unusual images and precisely locating the anomalies. It has been widely applied in various domains, such as quality control in industries. Due to the inherent data imbalance in AD, where abnormal images are far less common than normal ones, unsupervised methods have been explored. These methods learn exclusively from normal images of a single object class. Although such single-class unsupervised AD approaches~\cite{mvtec, vevae, itad, hou2021divide, tsdn, fapm, padim, patchcore} have shown strong performance, they struggle to work effectively on unseen object classes, making universal application challenging. This limitation has led to the emergence of few-shot AD~\cite{fewshot1, fewshot2, fewshot3}, multi-class AD~\cite{multi-class1,multi-class2,multi-class3} and zero-shot anomaly detection (ZSAD)~\cite{april-gan, anomalyclip, adaclip,  winclip} research. In particular, the demand for ZSAD, which can be widely applied to various datasets, has increased. ZSAD aims to train a model using a small auxiliary dataset and to enable it to detect anomalies in object classes not present in the training data. This capability presents substantial practical value in real-world scenarios where labeled anomaly data are unavailable.

Recently, vision-language models (VLMs)~\cite{align, clip, flava, blip, blip2}, such as CLIP~\cite{clip} have gained prominence as a foundational VLM for ZSAD, owing to their robust generalization capabilities. Trained contrastively on vast datasets of image-text pairs, VLMs are able to extract powerful features across diverse objects. WinCLIP~\cite{winclip} is a pioneering work that explored the potential of VLMs in ZSAD. As a train-free approach, it directly computed the similarity between CLIP vision and text features extracted from input image and numerous fixed text templates, respectively. 
However, since CLIP is not industry-focused setting, it is challenging to derive a generalized text embedding that effectively distinguishes normal and abnormal images across diverse categories.
Consequently, prompt learning-based methods~\cite{anomalyclip, adaclip} have been suggested to adapt CLIP prompt embeddings for ZSAD tasks. AnomalyCLIP~\cite{anomalyclip} (Fig.~\ref{fig:fig1}(a)) proposed a general object-agnostic prompt learning method that can be broadly applied across various domains. AdaCLIP~\cite{adaclip} proposed hybrid learnable prompts, combining static prompts and dynamic prompts generated from the final-layer token embedding of each image like Fig.~\ref{fig:fig1}(b). However, AnomalyCLIP faces the challenge of using very general query prompts, making it difficult to consider the class specific cues. On the other hand, AdaCLIP relied on dynamic prompts using class tokens. This approach revealed that static prompts often become entangled with confusing dynamic prompts during inference.

To address these challenges, we propose GenCLIP (Generalizing CLIP Prompts), a framework that focuses on improving the learning of the General Query Prompt tokens (GQPs), which plays a central role in both generalized normal/abnormal text embeddings and object class-aware text embeddings. The GQPs are learned by incorporating Multi-layer Vision Prompt tokens (MVPs) extracted from various layers of the CLIP vision encoder. Specifically, we design a learnable query prompt augmentation strategy that integrates these multi-layer vision features. As shown in Fig.~\ref{fig:fig1}(c), GenCLIP enhances both normal and abnormal text prompts, which share a common learnable query prompt, by incorporating information from multiple vision feature-based prompts. These vision feature-based prompts provide rich semantic context for the image-to-text matching process, effectively augmenting the GQPs in the feature dimension. By fusing the general prompts with various vision features and utilizing them for feature matching, the general prompts are naturally regularized, improving their stability and robustness. This approach enables the text prompts to learn from a broader range of visual information, from low-level textures and edges to high-level semantic concepts, resulting in richer and more comprehensive text embeddings for anomaly detection.

During inference, we employ a dual-branch strategy consisting of vision-enhanced and query-only branches. The first branch, the vision-enhanced branch, follows a process similar to training by combining multi-layer vision prompts with the query prompt. Since effectively incorporating image information into text prompts is crucial in this branch, we introduce a class name filtering (CNF) method to refine textual representations and improve alignment with image features. 

The second branch, the query-only branch, utilizes a highly generalized GQPs without incorporating vision features or class names. Instead of relying on explicit visual cues, GQPs are trained to encode general representations of normal and abnormal instances, making it adaptable across diverse object categories. The GQPs are utilized without incorporating vision features or class names, which helps identify outliers where class information is not beneficial for anomaly detection.

\vspace{1mm}
\noindent We highlight our major contributions as:
\begin{itemize}
    \item \textbf{Multi-layer prompting.} GenCLIP enhances text prompts by incorporating visual prompt tokens extracted from multi-layer visual features, helping to mitigate overfitting while effectively capturing class-specific information.
    
    \item \textbf{Novel inference strategy.} Our innovative inference strategy employs a dual-branch approach. The first branch, the vision-enhanced branch, enriches the query prompt with multi-layer visual feature prompts and CNF to improve semantic representation. The second branch, the query-only branch, uses a class-independent query prompt to capture general normal and abnormal patterns. This combined strategy boosts both the robustness and performance of GenCLIP.

    \item \textbf{SOTA results for ZSAD.} GenCLIP’s anomaly detection and segmentation capabilities are validated across six benchmark datasets. Our experimental results show that GenCLIP outperforms existing state-of-the-art methods.
    
\end{itemize}

\begin{figure*}[t]
\begin{center}
    \includegraphics[width=0.9\linewidth]{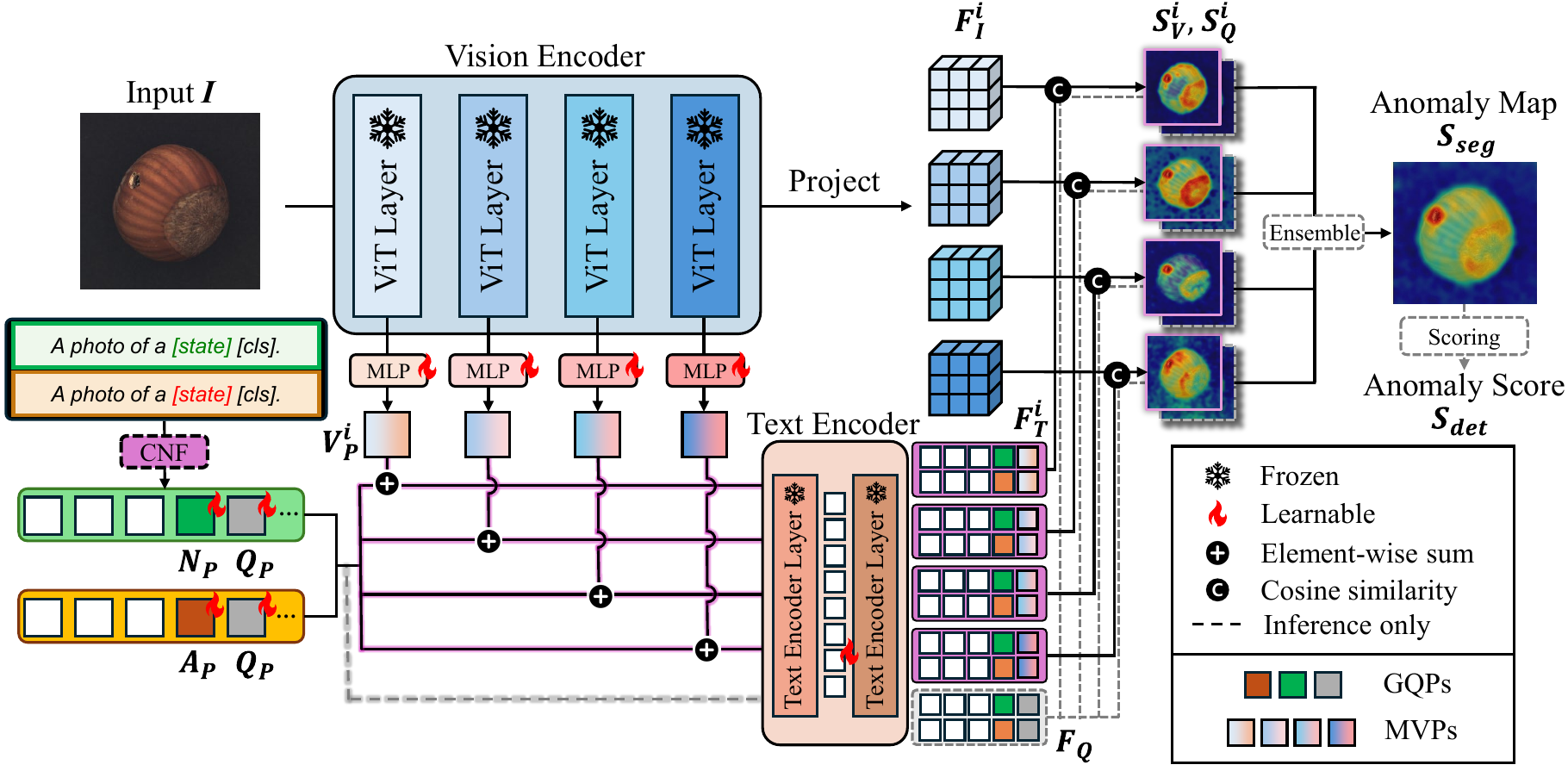}
\end{center}

\vspace{-4mm}

\definecolor{skyblue}{RGB}{135, 206, 235}

   \caption{The framework of GenCLIP. It consists of a CLIP vision encoder and a text encoder. The normal and abnormal prompts include individually learnable parameters $\mathbf{N_p}, \mathbf{A_p}$ and sharing learnable parameters $\mathbf{Q_P}$. During training, all modules except for CNF and $\mathbf{F_Q}$ are used. Given an image $\mathbf{I}$ and texts $\mathbf{T_N}$ and $\mathbf{T_A}$, GenCLIP outputs layer-wise score maps $\mathbf{S^i_V}$ by computing the similarity between the vision and text features. During inference, GenCLIP utilizes a two-branch inference strategy: \textcolor{violet}{Vision-enhanced branch} and \textcolor{gray}{Query-only branch} at the bottom of the figure. CNF is used only at the vision enhanced-branch. }
   \vspace{-0.5cm}

\label{fig2}
\end{figure*}
\section{Related Work}

\subsection{Zero-Shot Anomaly Detection}

The advent of vision-language models (VLMs) like CLIP, which demonstrate exceptional performance across a range of zero-shot tasks, has sparked significant interest in applying these models to zero-shot anomaly detection (ZSAD). In ZSAD, models are trained on auxiliary datasets to distinguish between normal and abnormal patterns, even in unseen categories.

WinCLIP~\cite{winclip} was the first to explore ZSAD using CLIP. It leveraged numerous normal and abnormal text templates to extract corresponding embeddings and compared them with patch features from the CLIP image encoder for anomaly detection. While WinCLIP achieved impressive results by capitalizing on CLIP's general capabilities, it was limited by the fact that CLIP was not specifically designed for anomaly detection. 

To address this, AnomalyCLIP~\cite{anomalyclip} introduced learnable query prompts to develop highly generalizable normal and abnormal embeddings, independent of object classes. Additionally, AnomalyCLIP adopted the term "object" instead of class names, providing a more general representation for all images. Building on this idea, AdaCLIP~\cite{adaclip} emphasized the importance of dynamic prompts tailored to individual images, proposing methods that adapt text prompts using image features to provide more fine-grained details in the text prompt. Moreover, VCP-CLIP~\cite{vcpclip}, which targets anomaly segmentation, enhanced text prompts by integrating visual context using VCP modules while adjusting text embeddings with image-specific information.

Many previous ZSAD methods~\cite{winclip, anomalyclip, adaclip} compute segmentation scores based on the similarity between patch features obtained through V-V attention~\cite{clipsurgery} in the CLIP vision encoder and text embeddings, using CLIP’s class token to derive detection scores. In contrast, GenCLIP retains the original CLIP vision encoder without V-V attention and derives detection scores directly from segmentation scores. Additionally, unlike previous methods that rely solely on either general prompts or class-specific prompts, GenCLIP proposes a balanced approach that effectively utilizes both.

\subsection{Prompt Learning}

Prompt learning methods~\cite{coop, cocoop, denseclip, HPT, coprompt} have become a key approach to adapting vision-language models (VLMs) for various downstream tasks, offering a more flexible and resource-efficient alternative to traditional fine-tuning. Unlike conventional fine-tuning, which requires updating the entire model’s parameters, prompt learning focuses on modifying the input prompts while maintaining the pre-trained model’s generalization capabilities. This paradigm has proven particularly effective in leveraging large-scale VLMs, such as CLIP~\cite{clip} and ALIGN~\cite{align}, for diverse applications.

Early works in prompt learning, such as CoOp (Context Optimization)~\cite{coop}, introduced learnable textual prompts to replace manually crafted ones. By optimizing continuous prompt vectors, CoOp enabled task-specific alignment between textual and visual features, leading to significant improvements in performance across downstream tasks. However, despite its success, prompt learning faces challenges when dealing with data outside its training distribution and struggles to adapt to changes in the target domain.

To address these issues, subsequent works like CoCoOp~\cite{cocoop} introduced context-conditioned prompts, which dynamically adapt to individual input images, thus enhancing generalization to unseen categories. Similarly, DenseCLIP~\cite{denseclip} extended prompt learning to dense prediction tasks by aligning dense visual features with textual embeddings, further demonstrating the versatility of prompt learning for domain adaptation and cross-modal understanding. 

\section{Methods}
\subsection{Overview}
In this paper, we propose \textbf{GenCLIP}, a robust and generalizable prompt learning framework for zero-shot anomaly detection (ZSAD) across various industrial datasets. 

Similar to prior ZSAD methods~\cite{winclip, anomalyclip, adaclip}, \textbf{GenCLIP} computes the similarity between an image and predefined normal/abnormal text descriptions, such as \textit{``A photo of a good [cls]''} and \textit{``A photo of a damaged [cls]''}, where \textit{[cls]} represents the target object class. As illustrated in Fig.~\ref{fig2}, the input image $\mathbf{I} \in \mathbb{R}^{h \times w \times3}$ is processed through the pre-trained CLIP vision encoder. Embeddings from multiple layers are passed through separate linear layers to generate vision prompt tokens $\mathbf{V^{\mathit{i}}_P}$, where $i$ denotes the layer index. $\mathbf{V^{\mathit{i}}_P}$ are then incorporated into the query prompt token $\mathbf{Q_P}$. The augmented prompts are subsequently transformed into text embeddings $\mathbf{F^{\mathit{i}}_T}$ via the CLIP text encoder. Finally, cosine similarity is computed between the layer-wise text embeddings $\mathbf{F^{\mathit{i}}_T}$ and vision features $\mathbf{F^{\mathit{i}}_I}$.

\subsection{Multi-Layer Prompting}
\label{mvp}

Previous ZSAD methods~\cite{adaclip, vcpclip} that leverage vision features for prompt learning have demonstrated strong performance owing to their vision adaptation capabilities. However, these approaches primarily rely on the class token from the CLIP image encoder, which limits the model’s ability to capture the full spectrum of visual information and increases the risk of overfitting to the small auxiliary dataset. In ZSAD, it is crucial not only to obtain vision-adaptive text embeddings but also to extract generalizable embeddings that can distinguish between normal and abnormal instances, regardless of object class.

To address this, we propose a multi-layer vision feature prompting strategy that integrates General Query Prompt tokens (GQPs) with Multi-layer Vision Prompt tokens (MVPs). By incorporating features from multiple layers, our approach prevents bias toward a single level of feature representation, enhancing generalization across diverse anomaly types.

\vspace{1mm}
\noindent \textbf{General Query Prompt Tokens.}~ 
We construct two distinct text prompts: one representing the ``normal'' state and the other representing the ``abnormal'' state. These two types of text prompts, denoted as $\mathbf{T} \in \mathbb{R}^{2\times N_L \times C_T}$, are structured as:
\begin{center}
\vspace{-1mm}
\texttt{$\mathbf{T} =$}\texttt{[$\mathbf{N_P},\mathbf{A_P}$][photo][of][a]} \\
\vspace{5pt} 
\texttt{[state][cls][$\mathbf{Q_P}$] ...}
\end{center}
\noindent where $C_T$ represents the dimensionality of the text feature channels and $N_L$ denotes the total number of text tokens. These structured text prompts follow a unified template, each containing a learnable state-specific prompt token: $\mathbf{N_P}, \mathbf{A_P} \in \mathbb{R}^{C_T}$ for normal and abnormal states, respectively. Additionally, both text prompts share a set of learnable query prompt tokens $\mathbf{Q_P} \in \mathbb{R}^{C_T}$ during training. These prompt tokens GQPs are trained to differentiate between normal and abnormal images in a class-independent manner, enabling ZSAD across various object categories.

\vspace{1mm}
\noindent \textbf{Multi-Layer Vision Prompt Tokens.}~ 
Through the frozen vision encoder of CLIP, we extract patch features $\mathbf{\widetilde{F}_V^{\mathit{i}}} \in \mathbb{R}^{(1+H \times W) \times C_i}, i\in \{1, 2, \ldots, L\}$ at each layer, where $L$ is the number of layers in the CLIP vision encoder, $C_i$ is the dimensionality of the vision embedding features and $H=h/patch~size, W= w/patch~size$. These image features are then used to generate vision prompt tokens MVPs as follows:
\begin{equation}
\mathbf{V^{\mathit{i}}_P} = \text{$MLP^{\mathit{i}}_T$}(\text{mean}(\mathbf{\widetilde{F}_V^{\mathit{i}}}))~, \quad \mathbf{V^{\mathit{i}}_P} \in \mathbb{R}^{C_T},
\end{equation}

\noindent where $MLP^{\mathit{i}}_T$ is a learnable single-layer linear projection. The layer-specific vision prompt tokens $\mathbf{V^{\mathit{i}}_P}$ are then added to the shared query prompt token $\mathbf{Q_P}$ for both the general normal and abnormal prompts:

\begin{equation}
\mathbf{V_Q^{\mathit{i}}}= \mathbf{Q_P}+ \mathbf{V^{\mathit{i}}_P}.
\end{equation}

\noindent These vision-tuned tokens $\mathbf{V_Q^{\mathit{i}}}$ capture relevant visual context, bridging the visual and textual domains. By generating these tokens, the model directly injects visual information into the text embedding process, enabling a more nuanced integration of visual cues within the language model's context. Since different vision-tuned tokens $\mathbf{V_Q^{\mathit{i}}}$ are added to $\mathbf{Q_P}$, the general learnable prompts are effectively augmented. This diverse vision feature-based prompt augmentation enhances the robustness of the GQPs, reducing the risk of overfitting.

\vspace{1mm}
\noindent \textbf{Text Layer Prompt Tuning.}~ 
Before being passed into the CLIP text encoder, the text prompt $\mathbf{T^{\mathit{i}}} \in \mathbb{R}^{2\times N_L \times C_T}$ is modified with $\mathbf{V_Q^{\mathit{i}}}$ as follows:

\begin{center}
\vspace{-1mm}
\texttt{$\mathbf{T^{\mathit{i}}_V} =$}\texttt{[$\mathbf{N_P},\mathbf{A_P}$][photo][of][a]}\\
\vspace{5pt} 
\texttt{[state][cls][$\mathbf{V_Q^{\mathit{i}}}$]}
\vspace{-1mm}
\end{center}

\noindent The modified text prompt $\mathbf{T^{\mathit{i}}_V}$ is then processed through the frozen CLIP text encoder, generating the final text embeddings. In the original CLIP text encoder, fixed token embeddings are used, and since it was primarily designed for text processing, introducing learnable prompt tokens can lead to misalignment between text and image features. To mitigate this issue, similar to previous methods~\cite{anomalyclip, adaclip, vcpclip}, GenCLIP inserts learnable tokens before each Transformer layer’s text embeddings, as illustrated in Fig.~\ref{fig2}. These learnable tokens refine the text embeddings, optimizing their alignment with the new prompts and improving the overall semantic consistency between text and image features. The normal and abnormal text embeddings produced by the CLIP text encoder through this process are denoted as $\mathbf{F^{\mathit{i}}_T} \in \mathbb{R}^{2\times C_T}$ for each respective $\mathbf{V_Q^{\mathit{i}}}$.

\subsection{Optimization}
For training, vision patch features from multiple layers of the CLIP image encoder are extracted and directly utilized to generate the anomaly score map. The vision features $\mathbf{\widetilde{F}_V^{\mathit{i}}}$, excluding the class token, is obtained from each layer and projected to $\mathbf{F_I^{\mathit{i}}}$ using a dedicated single linear layer $MLP^{\mathit{i}}_I$, specifically trained for that layer. The cosine similarity between the projected patch features $\mathbf{F^{\mathit{i}}_I}$ and the tuned text embeddings $\mathbf{F^{\mathit{i}}_T}$ for each layer is then computed, enabling the construction of layer-specific score maps that distinguish normal and abnormal categories.

\begin{equation}
\mathbf{F^{\mathit{i}}_I} = \text{$MLP^{\mathit{i}}_I$}(\mathbf{\widetilde{F}_V^{\mathit{i}}})~, \quad \mathbf{F^{\mathit{i}}_I} \in \mathbb{R}^{(H \times W) \times C_T}
\end{equation}

\begin{equation}
\mathbf{S^{\mathit{i}}_V} = \text{Softmax}(\text{Up}(\cos(\mathbf{F^{\mathit{i}}_I}, \mathbf{F^{\mathit{i}}_T})))
\end{equation}

\noindent where $\text{Up}(\cdot)$ denotes bilinear interpolation for upsampling the score maps, and $\cos(\cdot)$ represents the cosine similarity function. 

GenCLIP is optimized using focal loss~\cite{focal} and dice loss~\cite{dice}, computed between the generated score maps $\mathbf{S^{\mathit{i}}_V}$ and the binary ground truth map $\mathbf{GT_{map}}$. The total loss function is defined as follows:

\begin{equation}
L_{\text{total}} = \sum_{i=1}^{L} \left( \text{Focal}(\mathbf{GT_{map}},\mathbf{S^{\mathit{i}}_V}) + \text{Dice}(\mathbf{GT_{map}}, \mathbf{S^{\mathit{i}}_V}) \right)
\end{equation}
\subsection{Class Name Filtering}
\begin{figure}[t]
\vspace{-3mm}
\begin{center}
    \includegraphics[width=1.0\linewidth]{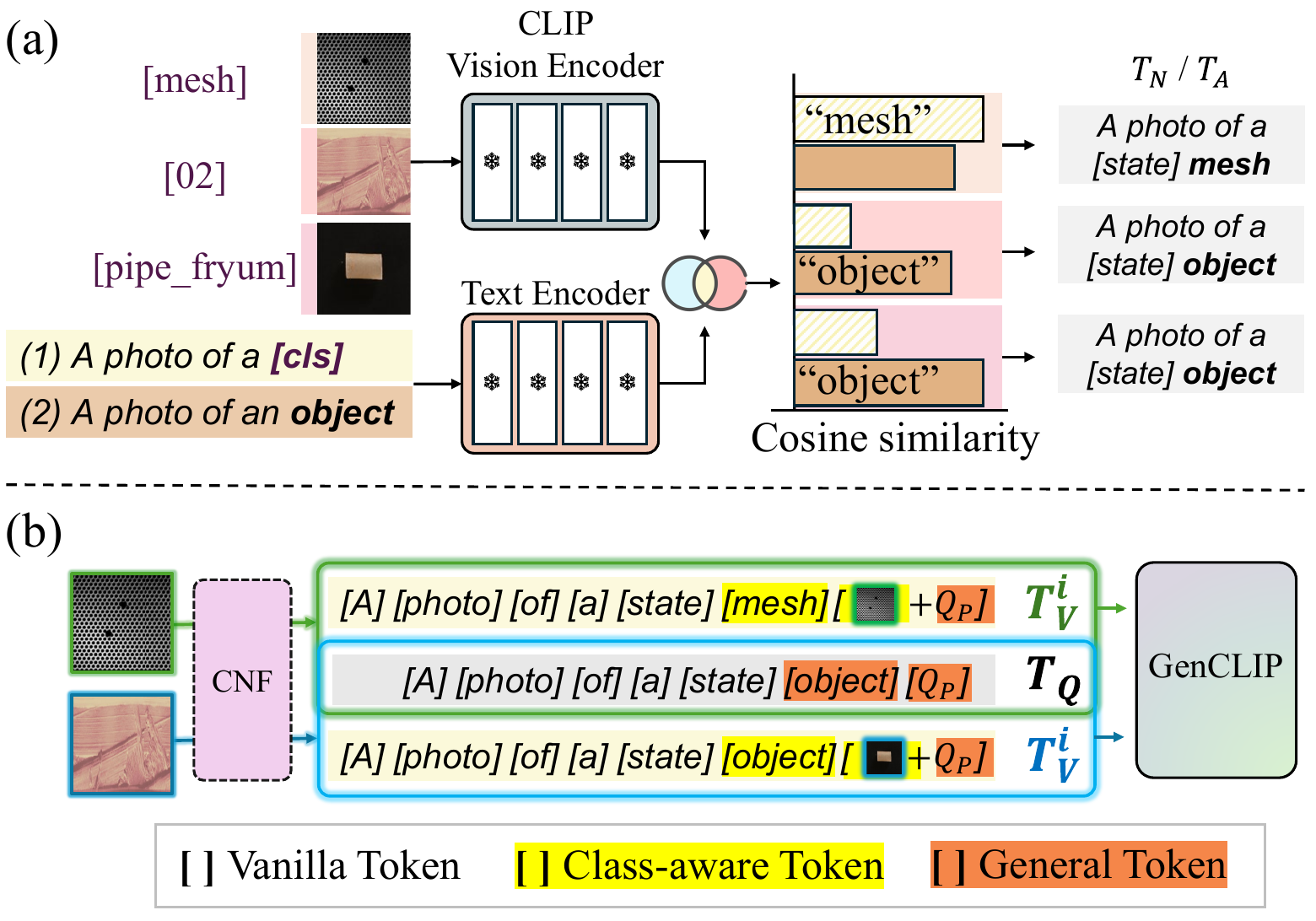}
\end{center}
   \caption{\textbf{(a) The architecture of CNF.} CNF utilizes the frozen CLIP image encoder and text encoder to replace ambiguous class names with the generic term ``object,'' eliminating unnecessary information and preventing potential confusion in the text encoder. \textbf{(b) Text prompts after CNF.} Class-aware text prompts $\mathbf{T^i_V}$ and the general text prompt $\mathbf{T_Q}$ are input to GenCLIP. $\mathbf{T_Q}$ is a unified prompt regardless of class.}
\vspace{-4mm}
\label{fig3}
\end{figure}

Our default input text follows the template: \textit{``A photo of a [state] [cls].''} In standard approaches, the ``\textit{[cls]}'' is typically replaced with the object class name specified in the dataset. This approach is effective when the class name accurately describes the image. However, in industrial applications, finding an exact textual description that properly represents an image is often challenging. Additionally, class names frequently fail to provide a clear semantic representation of the visual content. For instance, as shown in Fig.~\ref{fig3}(a), a wooden plank might be labeled as ``02,'' or a specific pipe component could be named ``\text{pipe\_fryum},'' both of which are uncommon textual representations. Since such labels do not clearly describe the object's appearance or function, they may lead to suboptimal feature representations, making it harder for the CLIP model to distinguish between normal and abnormal instances. Consequently, the use of ambiguous class names can introduce noise into the anomaly detection process.

To mitigate this issue, we propose Class Name Filtering (CNF), a method that refines textual representations during the vision-enhanced branch of inference. Before directly using the class name, CNF first removes numerical notations that do not contribute to meaningful image-class alignment. For example, ``pcb2'' is converted to ``pcb.'' Next, CNF leverages a frozen CLIP model to compute the cosine similarity between the target image and two different sentences: (1) a sentence using the original class name—\textit{``A photo of a [cls]''}—and (2) a sentence using a generic placeholder—\textit{``A photo of an object''} (Fig.~\ref{fig3}(a)). The term \textit{``object''} is chosen as a representative word encompassing all industrial images. If the target image exhibits a higher similarity to the generic term rather than the specific class name, CNF replaces the [cls] with \textit{``object.''} 

This adaptive text refinement strategy enhances the alignment between image and text embeddings, leading to more robust anomaly detection across diverse datasets. By replacing low-confidence class names with generic term \textit{``object,''} CNF ensures textual descriptions remain semantically meaningful within CLIP's learned representation space. This adjustment improves the compatibility between textual and visual features, allowing the CLIP text encoder to align better and interpret class-relevant information.

\begin{table*}[t!]
\centering
\caption{\textbf{Comparison of ZSAD methods in the industrial domain.} The best and the second best performance are highlighted \textcolor{red}{red}, and \textcolor{blue}{blue}, respectively. 
The \textdagger{} symbol indicates results obtained by re-training the model under the standard setting, as the original AdaCLIP paper did not provide values for these metrics.
The remaining results were taken from the AnomalyCLIP paper.}
\vspace{-1mm}
\label{tab:industrial}
\resizebox{0.95\linewidth}{!}{
\begin{tabular}{@{}c|c|cccccc@{}}
\toprule[1.5pt]
\rowcolor{gray!20} Metric & \cellcolor{gray!20}{Dataset} & CLIP~\cite{clip} & CoOp~\cite{coop} & WinCLIP~\cite{winclip} & AnomalyCLIP~\cite{anomalyclip} & \textdagger{}AdaCLIP~\cite{adaclip} & GenCLIP \\ 
\midrule
\multirow{7}{*}{\rotatebox{90}{\shortstack{Pixel-level\\(AUROC, PRO)}}}
 & MVTec         & (38.4, 11.3)& (33.3, 6.7) & (85.1, 64.6) & (\textcolor{blue}{91.1}, \textcolor{blue}{81.4}) & (90.2, 40.9) & (\textcolor{red}{92.7}, \textcolor{red}{88.1}) \\
 & VisA          & (46.6, 14.8)& (24.2, 3.8) & (79.6, 56.8) & (\textcolor{blue}{95.5}, \textcolor{blue}{87.0}) & (\textcolor{red}{96.0}, 73.0) & (95.3, \textcolor{red}{89.3}) \\
 & MPDD          & (62.1, 33.0)& (15.4, 2.3) & (76.4, 48.9) & (\textcolor{red}{96.5}, \textcolor{blue}{88.7}) & (95.7, 78.5) & (\textcolor{blue}{96.2}, \textcolor{red}{89.3}) \\
 & BTAD          & (30.6, 4.4) & (28.6, 3.8) & (72.7, 27.3) & (\textcolor{red}{94.2}, \textcolor{blue}{74.8}) & (89.5, 32.9) & (\textcolor{blue}{93.6}, \textcolor{red}{75.6}) \\
 & SDD           & (39.0, 8.9) & (28.9, 7.1) & (68.8, 24.2) & (90.6, \textcolor{blue}{67.8}) & (\textcolor{red}{97.4}, 56.0) & (\textcolor{blue}{96.8}, \textcolor{red}{94.9}) \\
 & DTD-Synthetic & (33.9, 12.5)& (-, -)      & (83.9, 57.8) & (\textcolor{blue}{97.9}, \textcolor{blue}{92.3}) & (\textcolor{red}{98.6}, 82.0) & (\textcolor{blue}{97.9}, \textcolor{red}{93.6}) \\ 
 \cmidrule(l){2-8}
 & Average       & (41.8, 14.2)& (26.1, 4.7) & (77.8, 46.6) & (94.3, \textcolor{blue}{82.0}) & (\textcolor{blue}{94.6}, 61.0) & (\textcolor{red}{95.4}, \textcolor{red}{88.5}) \\ 
 \midrule
\multirow{7}{*}{\rotatebox{90}{\shortstack{Image-level\\(AUROC, AP)}}}
 & MVTec         & (74.1, 87.6) & (88.8, 94.8) & (\textcolor{red}{91.8}, \textcolor{red}{96.5}) & (\textcolor{blue}{91.5}, \textcolor{blue}{96.2}) & (91.3, 96.1) & (90.9, 96.1) \\
 & VisA          & (66.4, 71.5) & (62.8, 68.8) & (78.1, 81.2) & (\textcolor{blue}{82.1}, \textcolor{blue}{85.4}) & (75.9, 78.3) & (\textcolor{red}{83.3}, \textcolor{red}{87.5}) \\
 & MPDD          & (54.3, 65.4) & (55.1, 64.2) & (63.6, 69.9) & (\textcolor{red}{77.0}, \textcolor{red}{82.0}) & (72.7, \textcolor{blue}{81.0})   & (\textcolor{blue}{73.7}, 79.6) \\
 & BTAD          & (34.5, 52.5) & (66.8, 77.4) & (68.2, 70.9) & (\textcolor{blue}{88.3}, 87.3) & (83.8, \textcolor{blue}{89.5}) & (\textcolor{red}{90.0}, \textcolor{red}{96.9}) \\
 & SDD           &  (65.7, 45.2)& (74.9, 65.1) & (84.3, 77.4) & (84.7, \textcolor{blue}{80.0}) & (\textcolor{red}{96.8}, \textcolor{red}{84.1}) & (\textcolor{blue}{92.2}, 68.9) \\
 & DTD-Synthetic & (71.6, 85.7) & (-, -)       & (\textcolor{blue}{93.2}, 92.6) & (\textcolor{red}{93.5}, \textcolor{red}{97.0}) & (91.9, 95.8) & (88.8, \textcolor{blue}{96.3}) \\ 
 \cmidrule(l){2-8}
  & Average      & (61.1, 68.0) & (69.7, 74.1) & (79.9, 81.4) & (\textcolor{blue}{86.2}, \textcolor{red}{88.0}) & (85.4, 87.5) & (\textcolor{red}{87.3}, \textcolor{blue}{87.6}) \\ 
\bottomrule[1.5pt]
\end{tabular}
}
\end{table*}
\vspace{-2mm}

\subsection{Inference}
\label{sub_inference}

\noindent \textbf{Vision-enhanced branch.} The vision-enhanced branch follows a process similar to training. Given an input image $\mathbf{I}$, we extract patch-level vision feature maps $\mathbf{\widetilde{F}_V^{\mathit{i}}}, {\mathit{i}} \in \{1,2,\cdots, L\}$ from multiple layers of the frozen CLIP image encoder. These patch features are then processed through $MLP_I^{\mathit{i}}$ and $MLP_T^{\mathit{i}}$, transforming into $\mathbf{F_I^{\mathit{i}}}$, which is aligned with the text embedding dimension, and $\mathbf{V_P^{\mathit{i}}}$, which encodes class and layer-specific information for the text prompt. Unlike during training, the template sentences are processed through CNF to adaptively align image and text representations. 

The generated text prompts, after passing through the tokenizer, are supplemented with the learnable prompt tokens $\mathbf{N_P}, \mathbf{A_P}$, and $\mathbf{Q_P}$. These GQPs are further augmented with multi-layer vision prompt tokens $\mathbf{V_P^{\mathit{i}}}$, generating $L$ distinct image-layer semantic text prompts $\mathbf{F_T^{\mathit{i}}}$. The final anomaly score map in the vision-enhanced branch is computed as follows:

\begin{equation}
	\label{eq:anomaly_score_map}
	\mathbf{S^{\mathit{i}}_V} = \text{Softmax}\left(\text{Up}(\cos(\mathbf{F_I^{\mathit{i}}}, \mathbf{F_T^{\mathit{i}}})) \right)
\end{equation}

\vspace{1mm}
\noindent \textbf{Query-only branch.}~
For ZSAD, it is crucial not only to obtain text embeddings that align with a given image but also to learn generalized text embeddings that effectively represent normal and abnormal states across diverse object categories. During training, GQPs undergo augmentation with diverse vision information, allowing the model to learn generalizable representations of normal and abnormal concepts. 

At inference time, we explicitly use a general text embedding $\mathbf{T_Q}$ based solely on the query prompt. Additionally, instead of using the actual class name, we employ the universal class term ``object,'' ensuring robust and adaptable anomaly detection. As illustrated in Fig.~\ref{fig3}(b), the term ``object'' in $\mathbf{T_Q}$ serves as a highly general and broadly applicable descriptor, distinct from the ``object'' used in prompts where CNF has replaced the class label.

\begin{center}
\texttt{$\mathbf{T_Q} =$}\texttt{[$\mathbf{N_P}, \mathbf{A_P}$][photo][of][a][state]}\\
\vspace{2pt} 
\texttt{[object][$\mathbf{Q_P}$]}
\end{center}

The highly class-agnostic text prompt $\mathbf{T_Q}$ serves as a fixed prompt that consistently represents normal and abnormal states, remaining unchanged across different datasets, images, and object classes. It is transformed into a class-agnostic text embedding $\mathbf{F_Q} \in \mathbb{R}^{{2\times C_T}}$ via the CLIP text encoder, as illustrated in Fig.~\ref{fig2}. The image patch feature $\mathbf{F_I^{\mathit{i}}}$, which is used to extract the score map, is the same as that in the vision-enhanced branch. The final score maps of the query-only branch are computed as follows:

\begin{equation}
	\label{eq:anomaly_score_map}
	\mathbf{S^{\mathit{i}}_Q} = \text{Softmax}\left(\text{Up}(\cos(\mathbf{F_I^{\mathit{i}}}, \mathbf{F_Q})) \right)
\end{equation}

\vspace{1mm}
\noindent \textbf{Pixel-level Anomaly Segmentation.}~ The anomaly score maps $\mathbf{S_V^{\mathit{i}}}$, obtained through the vision-enhanced branch, capture detailed texture and semantic information. In contrast, $\mathbf{S_Q^{\mathit{i}}}$ are anomaly maps derived from the query-only branch, reflecting anomalies from a general object perspective. To improve the robustness of anomaly localization, we combine these anomaly score maps through a weighted summation:
\vspace{-1mm}
\begin{equation}
	\mathbf{S_{seg}} = G\left(\alpha \sum_{i=1}^{L} \mathbf{S_V^{\mathit{i}}} + (1-\alpha) \sum_{i=1}^{L} \mathbf{S_Q^{\mathit{i}}}, \right)
    \label{eq9}
\end{equation}

\noindent where $\alpha$ is a weighting factor that controls the contribution of $\mathbf{S_V^{\mathit{i}}}$ and $\mathbf{S_Q^{\mathit{i}}}$, and $G(\cdot)$ represents Gaussian smoothing. The resulting score map, $\mathbf{S_{seg}}$, is used as the final map for anomaly segmentation.

\vspace{1mm}
\noindent \textbf{Image-level Anomaly Detection.} 
Many previous ZSAD methods~\cite{winclip, anomalyclip, adaclip} rely on the class token obtained from the CLIP vision encoder to perform anomaly detection inference. However, instead of using class tokens, we propose a method for image-level anomaly detection by taking the maximum value of the score map. The image-level anomaly score $\mathbf{S_{det}}$ is calculated using the following formulas:

\begin{equation}
	 w = \frac{\text{mean}\left(\exp{(\mathbf{S^{N_1}_{seg}})}\right)}{\text{mean}\left(\exp{(\mathbf{S^{N_2}_{seg}})}\right)}~~, N_1 < N_2
\end{equation}

\begin{equation}
	\mathbf{S_{det}} = w \cdot \text{mean}(\mathbf{S^{N_1}_{seg}})
\end{equation}

\noindent where $\mathbf{S^{N}_{seg}}$ represents the $N$ pixels with the highest anomaly scores in $\mathbf{S_{seg}}$. We leverage a confidence-based scoring method, enabling robust anomaly detection performance using only the score map.

\begin{figure*}[t]
\begin{center}
    \includegraphics[width=0.9\linewidth]{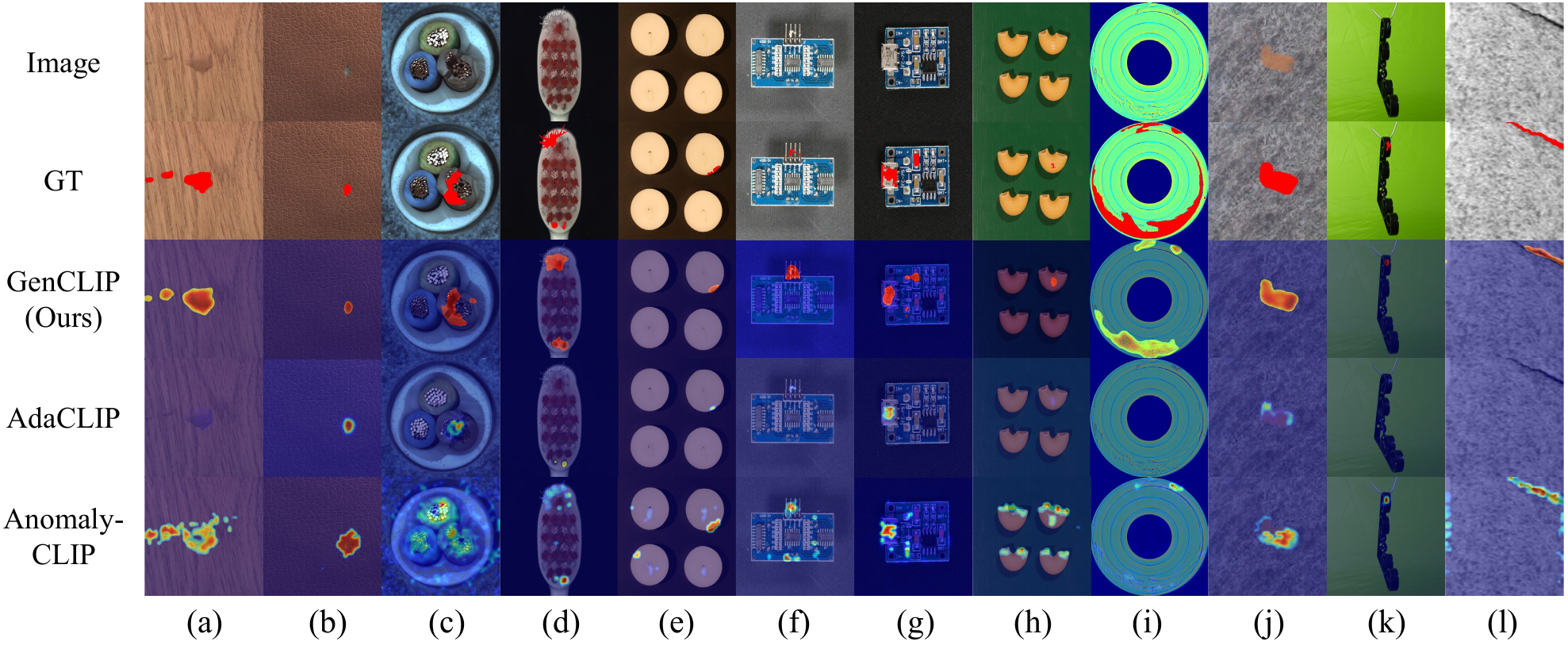}
\end{center}
\vspace{-5mm}
   \caption{Qualitative comparison against Adaclip and AnomalyCLIP on our representative datasets.}
\label{fig:quali}
\end{figure*}

\section{Experiments}
\subsection{Experimental Setup}

\noindent \textbf{Datasets.} To validate the ZSAD performance of GenCLIP, we evaluate it on six industrial ZSAD benchmark datasets: MVTec-AD~\cite{mvtec}, VisA~\cite{visa}, MPDD~\cite{MPDD}, BTAD~\cite{btech}, SDD~\cite{SDD}, and DTD-synthetic~\cite{dtd}. For a fair comparison, we follow the settings of existing works~\cite{winclip, adaclip, anomalyclip}. Specifically, we train GenCLIP using MVTec-AD when evaluating its performance on other datasets. For MVTec-specific evaluation, we use VisA as the training set. Please refer to Appendix~\ref{appendix:dataset} for a detailed description of all datasets.

\vspace{1mm}
\noindent \textbf{Evaluation Metrics.} To ensure consistency with recent studies~\cite{adaclip, winclip, anomalyclip}, we adopt the area under the receiver operating characteristic curve (AUROC), per-region overlap (PRO), and average precision (AP) as our primary evaluation metrics. These metrics provide a standardized and fair comparison with existing methods. 

\vspace{1mm}
\noindent \textbf{Implementation Details.} We utilize the CLIP model with the ViT-L-14-336 architecture, pre-trained by OpenAI~\cite{clip}. Patch-level features are extracted from four evenly spaced layers ($6$, $12$, $18$, and $24$) of the $24$-layer Transformer-based image encoder. Input images are resized to $518 \times 518$ pixels before processing. 

GenCLIP is trained using simple text templates: \textit{``A photo of a good [cls] object.''} and \textit{``A photo of a damaged [cls] object.''} We use a single instance of $\mathbf{N_P}$ and $\mathbf{A_P}$ each, along with two instances of $\mathbf{Q_P}$. Additional implementation details are provided in Appendix~\ref{appendix:implementation}.
s are conducted using PyTorch on a single Nvidia RTX A5000 GPU.

\begin{figure}[t]
\begin{center}
    \includegraphics[width=1.0\linewidth]{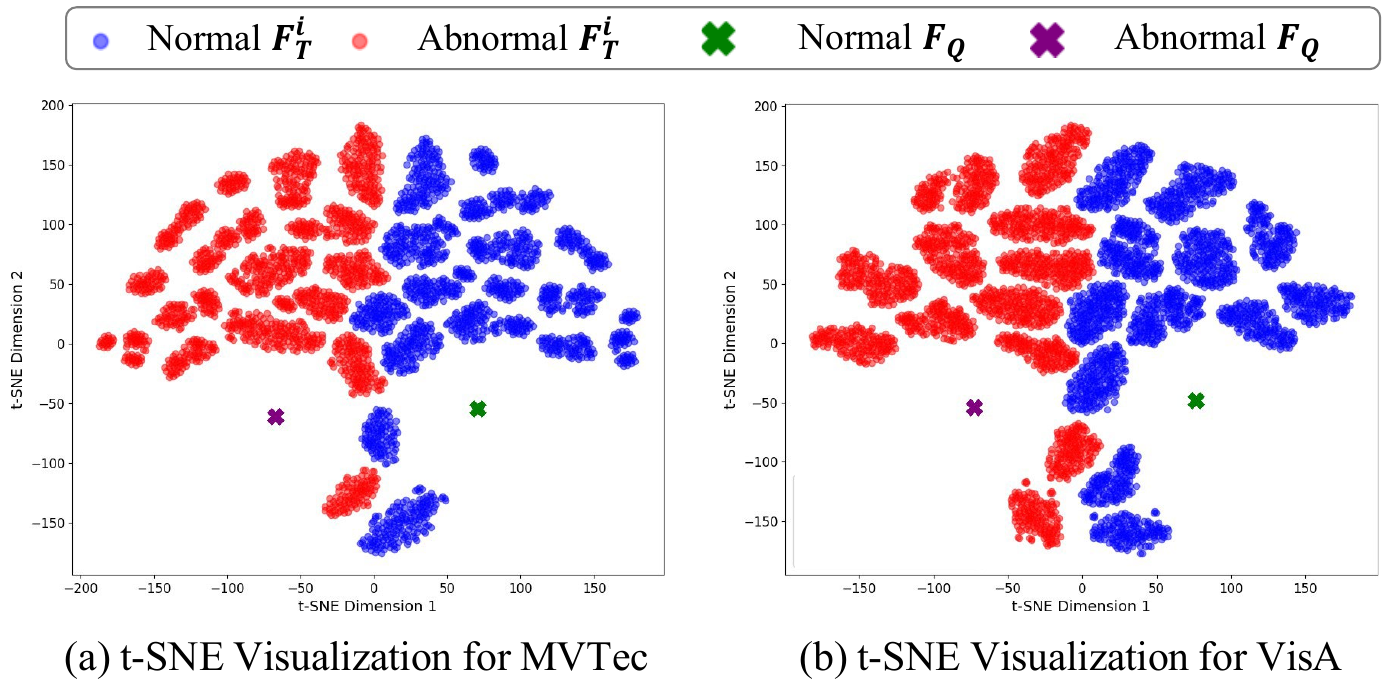}
\end{center}
\vspace{-0.4cm}
   \caption{t-SNE visualization of text features $\mathbf{F_T^{\mathit{i}}}$ and $\mathbf{F_Q}$ for (a) MVTec and (b) VisA test datasets.}
\label{fig:tsne}
\vspace{-0.2cm}
\end{figure}
\vspace{-1mm}

\begin{table}[]
\caption{Ablation study results of vision-enhanced branch (use multi-layer vision prompt tokens) and query-only branch (use only general query prompts) on VisA dataset. The best results are \textcolor{red}{red}, and the second-best are \textcolor{blue}{blue}.}
\label{tab2}
\centering
\resizebox{0.8\columnwidth}{!}{%
\begin{tabular}{cc|cc|cc}
\hline
\multirow{2}{*}{MVP} & \multirow{2}{*}{GQP}  & \multicolumn{2}{c|}{Pixel-level} & \multicolumn{2}{c}{Image-level}  \\ \cline{3-6} 
                     &                      & \multicolumn{1}{c|}{AUROC} & PRO & \multicolumn{1}{c|}{AUROC} & AP   \\ \hline
                     &                      & \multicolumn{1}{c|}{94.8}  & 87.9  & \multicolumn{1}{c|}{81.6}  & 86.4 \\ \cline{3-6} 
\checkmark           &                      & \multicolumn{1}{c|}{\textcolor{blue}{95.2}}  & \textcolor{blue}{88.7}  & \multicolumn{1}{c|}{\textcolor{blue}{82.2}}  & \textcolor{blue}{86.8} \\ \cline{3-6} 
\checkmark           & \checkmark           & \multicolumn{1}{c|}{\textcolor{red}{95.3}} & \textcolor{red}{89.3} & \multicolumn{1}{c|}{\textcolor{red}{83.3}} & \textcolor{red}{87.5} \\ \hline
\end{tabular}%
}
\end{table}

\begin{table}[]
\caption{Ablation studies on representative classes of VisA where Class name filtering (CNF) is activated. Cases where applying CNF improved performance are \textbf{\textcolor{red}{red}}.}
\label{tab:table3}
\centering
\resizebox{\columnwidth}{!}{%
\begin{tabular}{c|c|c|c|c}
\hline
\multirow{2}{*}{Class} & \multicolumn{2}{c|}{Pixel-level (AUROC, PRO)} & \multicolumn{2}{c}{Image-level (AUROC, AP)} \\ \cline{2-5} 
                       & w/ CNF  & w/o CNF  & w/ CNF  & w/o CNF \\ \hline
chewinggum  & \textcolor{black}{(}99.2 \textcolor{black}{,} 85.5\textcolor{black}{)}  & \textcolor{black}{(}99.2 \textcolor{black}{,} 85.5\textcolor{black}{)}  & {\color[HTML]{FE0000} \textcolor{black}{(}95.8 \textcolor{black}{,} 98.3\textcolor{black}{)}} & \textcolor{black}{(}95.6 \textcolor{black}{,} 98.2\textcolor{black}{)}  \\  
fryum       & {\color[HTML]{FE0000} \textcolor{black}{(}94.7 \textcolor{black}{,} 90.6\textcolor{black}{)}} & \textcolor{black}{(}94.5 \textcolor{black}{,} 89.3\textcolor{black}{)}  & {\color[HTML]{FE0000} \textcolor{black}{(}81.2 \textcolor{black}{,} 90.8\textcolor{black}{)}} & \textcolor{black}{(}77.5 \textcolor{black}{,} 88.8\textcolor{black}{)}  \\  
macaroni1   & {\color[HTML]{FE0000} \textcolor{black}{(}99.2 \textcolor{black}{,} 96.4\textcolor{black}{)}} & \textcolor{black}{(}99.1 \textcolor{black}{,} 95.9\textcolor{black}{)}  & {\color[HTML]{FE0000} \textcolor{black}{(}87.3 \textcolor{black}{,} 89.0\textcolor{black}{)}} & \textcolor{black}{(}86.6 \textcolor{black}{,} 88.8\textcolor{black}{)}  \\  
pcb1        & {\color[HTML]{FE0000} \textcolor{black}{(}91.9 \textcolor{black}{,} \textcolor{black}{88.7)}} & \textcolor{black}{(}91.8 \textcolor{black}{,} 89.0\textcolor{black}{)}  & {\color[HTML]{FE0000} \textcolor{black}{(}67.9 \textcolor{black}{,} 74.2\textcolor{black}{)}} & \textcolor{black}{(}63.6 \textcolor{black}{,} 71.5\textcolor{black}{)}  \\  
pipe\_fryum & {\textcolor{black}{(}96.2 \textcolor{black}{,} \color[HTML]{FE0000} {96.3\textcolor{black}{)}}} & \textcolor{black}{(}96.4 \textcolor{black}{,} 95.5\textcolor{black}{)}  & {\color[HTML]{FE0000} \textcolor{black}{(}90.5 \textcolor{black}{,} 95.4\textcolor{black}{)}} & \textcolor{black}{(}87.4 \textcolor{black}{,} 94.0\textcolor{black}{)}  \\  
\hline
mean of VisA  & (95.3, \textcolor{red}{89.3}) & (95.3, 89.0) & (\textcolor{red}{83.3}, \textcolor{red}{87.5}) & (82.4, 87.0) \\
\hline
\end{tabular}%
}
\vspace{-0.3cm}
\end{table}

\subsection{Quantitative Comparison}
We compare GenCLIP with existing ZSAD methods under the same task setting, including AdaCLIP~\cite{adaclip}, AnomalyCLIP~\cite{anomalyclip}, and WinCLIP~\cite{winclip}. Additionally, we include ZSAD performance results from representative VLMs such as CLIP~\cite{clip} and CoOp~\cite{coop} for further comparison.\\
\indent As shown in Tab.~\ref{tab:industrial}, GenCLIP achieves superior accuracy in both detecting and segmenting anomalies within industrial datasets, surpassing the compared methods. Since our model is trained solely using the anomaly score map, it exhibits particularly strong performance in pixel-level evaluation metrics. Specifically, GenCLIP significantly surpasses the second-best approaches, achieving a $0.8$\% point gain in average pixel-level AUROC and an exceptional $6.5$\% point improvement in PRO. Additionally, although our method shows a slight 0.4\% point drop in image-level AP, it achieves a notable 1.1\% point improvement in AUROC. For class-specific performance, please refer to Appendix~\ref{appendix:performance}.

\subsection{Qualitative Comparison}
In Fig.~\ref{fig:quali}, we compare our anomaly segmentation results against those of AdaCLIP and AnomalyCLIP. The results from different industrial datasets are presented as follows: MVTec-AD: (a)-(d), VisA: (e)-(h), BTAD: (i), DTD-synthetic: (j), MPDD: (k), and SDD: (l). GenCLIP successfully identifies and highlights defective regions with minimal false positives. The detected heatmaps closely align with the ground truth masks, whereas other methods exhibit less precise localization, often capturing unnecessary background noise. 
For instance, in (c), the anomaly on the cable is clearly delineated in GenCLIP’s result, whereas the competing methods produce scattered activations. Moreover, even small anomalies, such as the defects in (b), (e), and (h), are effectively detected, demonstrating the fine-grained sensitivity of our approach. GenCLIP successfully detects these structural inconsistencies with high precision, highlighting the superior capability of our method. Further results across all datasets are found in Appendix~\ref{appendix:dataset}. 

\subsection{Ablation Studies}
To demonstrate the effectiveness of our proposed method, we conduct ablation studies on the representative dataset VisA, as it contains a balanced distribution of both general anomalies (e.g., scratches, holes) and class-specific anomalies (e.g., missing parts, shape irregularities). 

\vspace{1mm}
\noindent \textbf{Effects of each branch.}~ As described in Sections~\ref{mvp} and~\ref{sub_inference}, $\mathbf{T^{\mathit{i}}_V}$ is not only class-specific but also tailored to the features extracted from different layers of the vision encoder. In contrast, $\mathbf{T_Q}$ is a class-agnostic text prompt, independent of both the dataset and object class. 

As shown in the t-SNE visualization in Fig.~\ref{fig:tsne}, the normal and abnormal embeddings $\mathbf{F^{\mathit{i}}_T}$ exhibit well-defined clustering due to the incorporation of class-specific and layer-specific information. Conversely, the normal and abnormal embeddings $\mathbf{F_Q}$ are positioned farther from these $\mathbf{F^{\mathit{i}}_T}$ clusters. However, they occupy representative locations that effectively capture the general concept of normality and abnormality. 

\noindent \textbf{Effects of MVP and GQP.}~ Results in Tab.~\ref{tab2} demonstrate the impact of MVPs and GQPs. Applying MVPs enhances anomaly detection performance compared to training with the query prompts and final-layer vision token, similar to AdaCLIP\cite{adaclip}. Additionally, image-AUROC and AP both show improvements, demonstrating the effectiveness of our approach. When applying query-only branch at inference time, we observe further enhancements across both pixel-level and image-level anomaly detection. Notably, the use of $\mathbf{F_Q}$ leads to better generalization, improving overall performance in detecting both fine-grained and structural anomalies. Please refer to further experiments and analyses provided in Appendix~\ref{appendix:layer} and \ref{appendix:generaltsne}

\vspace{1mm}
\noindent \textbf{Impacts of CNF.}~ Results in Tab.~\ref{tab:table3} demonstrate the effectiveness of CNF. Performance for representative classes in the VisA dataset, where CNF is activated, is reported. For cases such as ``chewinggum,'' ``fryum,'' and ``pipe\_fryum,'' class names were replaced with the more descriptive term ``object'' based on similarity score comparisons within the CLIP network, improving textual representations of the images. Additionally, numeric identifiers in class names like ``macaroni1'' and ``pcb1'' were removed, simplifying them to ``macaroni'' and ``pcb.'' As shown in the table, most performance metrics improved for classes where ``object'' replaced ambiguous class names, and numeric identifier removals led to overall class-wise performance gains. Consequently, we observed an overall improvement in the average performance on the VisA dataset. These results confirm that CNF enhances the effectiveness of the vision-enhanced branch, leading to a more robust anomaly detection performance. Furthermore, in Appendix~\ref{appendix:cnftext}, we report the results of various experiments conducted with different general terms.

\vspace{-2mm}
\section{Conclusion}
In this paper, we introduced GenCLIP, a novel approach for zero-shot anomaly detection (ZSAD) that leverages multi-layer visual feature prompts to enhance generalization and mitigate overfitting. Additionally, our dual-branch inference strategy enables robust anomaly detection across diverse domains. By integrating visual and textual cues effectively, GenCLIP provides a flexible and scalable solution for ZSAD.

\newpage
{
    \small
    \bibliographystyle{ieeenat_fullname}
    \bibliography{main}
}
\newpage

\appendix
\renewcommand{\thesection}{\Alph{section}} 

\section{Appendix}
In this appendix, we provide additional details on the datasets, supplementary experimental results, further analyses, an extended presentation of both quantitative and qualitative findings, and some failure cases. We use the same reference numbers with the main paper.

\section{Details}

\subsection{Dataset Details}~\label{appendix:dataset}
In this study, we conduct extensive experiments on six benchmark datasets covering various industrial classes. \\

\noindent \textbf{MVTec-AD~\cite{mvtec}.}~ MVTec-AD is a widely used industrial anomaly detection dataset, containing 5,354 color images with resolutions ranging from 700 to 1024 pixels. It includes 15 product categories, spanning both object (\textit{e.g.}, bottle, cable, capsule, metal nut, pill, screw, toothbrush, transistor, zipper) and texture (\textit{e.g.}, carpet, grid, leather, tile, wood) types, with pixel-level annotations. The dataset covers a diverse range of anomaly types, including missing components, deformations, contamination, scratches, cracks, and misalignment.

\noindent \textbf{VisA~\cite{visa}.}~ The VisA dataset consists of 10,821 high-resolution images (960 to 1500 pixels) designed for industrial anomaly detection. It covers 12 different object categories, including candle, capsule, cashew, chewing gum, frying pan, macadamia, PCB1, PCB2, pipe fryum, rope, rubber eraser, and screw, all annotated at the pixel level. The anomalies in this dataset include surface defects, dents, missing parts, deformation, and contamination.

\noindent \textbf{MPDD~\cite{MPDD}.}~ The Magnetic Particle Defect Dataset (MPDD) is specifically designed for detecting surface defects in metallic components using magnetic particle testing. It contains images capturing a variety of defect types, including cracks, pores, inclusions, and material fractures, with pixel-level annotations for segmentation-based anomaly detection tasks.

\noindent \textbf{BTAD~\cite{btech}.}~ BTAD is an anomaly detection dataset with three sub-datasets. Product 1 contains 400 images at a resolution of 1600 × 1600 pixels, Product 2 contains 1,000 images at 600 × 600 pixels, and Product 3 contains 399 images at 800 × 600 pixels. Each sub-dataset consists of images of different manufactured items, with common defect types such as scratches, deformations, misalignment, and material defects.

\noindent \textbf{SDD~\cite{SDD}.}~ The SDD dataset was collected in a controlled environment and comprises images of size 230 × 630 pixels. It contains 2,085 negative and 246 positive samples for training, along with 894 negative and 110 positive samples for testing. The anomalies in this dataset range from minor scratches and cracks to discoloration and material degradation, providing fine-grained segmentation masks for defect localization.

\noindent \textbf{DTD-Synthetic~\cite{dtd}.}~ The Describable Textures Dataset (DTD) consists of 5,640 images spanning 47 texture categories, such as \textit{striped, dotted, woven, bumpy, cracked, fibrous, fuzzy, knitted, perforated, scaly, smooth, wrinkled}, and more. It is widely used for texture classification tasks and serves as a benchmark for evaluating the generalization ability of models in recognizing diverse material textures. Anomalies in this dataset often involve unnatural texture patterns, structural irregularities, and deviations from expected texture distributions.

\subsection{Evaluation Metric Details}
To comprehensively evaluate the performance of our anomaly detection model, we employ four widely used metrics: Pixel-level AUROC, Pixel-level PRO, Image-level AUROC, and Image-level AP. These metrics assess the model's capability in both pixel-wise and image-wise anomaly detection, providing a holistic understanding of its effectiveness.

\noindent \textbf{Pixel-level AUROC.}~ The Area Under the Receiver Operating Characteristic Curve (AUROC) at the pixel level measures the model’s ability to distinguish between anomalous and normal pixels across varying threshold values. It is calculated based on the True Positive Rate (TPR) and False Positive Rate (FPR), defined as:
\begin{equation}
    TPR = \frac{TP}{TP + FN}, \quad FPR = \frac{FP}{FP + TN},
\end{equation}
\noindent where \( TP \) and \( FP \) represent the correctly and incorrectly classified anomalous pixels, while \( TN \) and \( FN \) denote the correctly and incorrectly classified normal pixels, respectively. A higher AUROC value indicates that the model can effectively separate anomalous pixels from normal ones, making it a crucial metric for pixel-wise anomaly localization.

\noindent \textbf{Pixel-level PRO.}~ The Area Under the Per-Region Overlap (PRO) is a metric designed to evaluate pixel-wise anomaly localization while considering the spatial consistency of anomalies. Instead of treating each pixel independently, PRO accounts for how well the predicted anomaly regions overlap with the ground truth anomaly regions. It is computed by integrating the per-region overlap (PRO) scores across multiple threshold levels. The PRO score at a given threshold is defined as:
\begin{equation}
    PRO = \frac{|P \cap G|}{|G|},
\end{equation}
\noindent where \( P \) is the set of predicted anomalous pixels and \( G \) is the ground truth anomaly region. A higher PRO indicates better spatial consistency in anomaly localization, making it a more robust measure than AUROC in scenarios where precise localization is essential.

\noindent \textbf{Image-level AUROC.}~ At the image level, AUROC evaluates the model’s ability to distinguish between anomalous and normal images. It follows the same definition as pixel-level AUROC but considers entire images instead of individual pixels. This metric is particularly useful for image-wise anomaly detection, where the goal is to determine whether an image contains any anomalies. A high image-level AUROC suggests that the model effectively discriminates between normal and abnormal images across various threshold settings.

\begin{table}[t]
\centering
\caption{Per-class performance on the MVTec dataset.}
\label{perclass_mvtec_results}
\resizebox{1\linewidth}{!}{
\begin{tabular}{@{}c|cc|cc@{}}
\toprule[1.5pt]
\rowcolor{gray!20} \multirow{2}{*}{Objects} & \multicolumn{2}{c|}{Pixel-level} & \multicolumn{2}{c}{Image-level} \\ 
\cmidrule(lr){2-3} \cmidrule(lr){4-5}
\rowcolor{gray!20} & AUROC (\%) & PRO (\%) & AUROC (\%) & AP (\%) \\
\midrule
Bottle     & 94.67 & 88.58 & 96.03 & 98.75 \\
Cable      & 81.20 & 74.98 & 83.06 & 90.86 \\
Capsule    & 96.98 & 94.69 & 93.54 & 98.68 \\
Carpet     & 99.66 & 98.65 & 100.00 & 100.00 \\
Grid       & 98.67 & 94.01 & 99.92 & 99.97 \\
Hazelnut   & 97.70 & 83.49 & 93.46 & 96.78 \\
Leather    & 99.60 & 99.01 & 100.00 & 100.00 \\
Metal Nut  & 76.50 & 72.74 & 66.72 & 91.50 \\
Pill       & 88.18 & 94.42 & 84.34 & 97.00 \\
Screw      & 98.64 & 93.26 & 89.18 & 95.75 \\
Tile       & 96.39 & 93.45 & 97.26 & 99.12 \\
Toothbrush & 94.98 & 88.85 & 85.28 & 94.12 \\
Transistor & 70.88 & 57.50 & 80.88 & 81.31 \\
Wood       & 97.53 & 94.88 & 95.26 & 98.50 \\
Zipper     & 98.32 & 93.10 & 98.77 & 99.71 \\
\midrule
Mean       & 92.66 & 88.10 & 90.91 & 96.14 \\
\bottomrule[1.5pt]
\end{tabular}
}
\end{table}

\begin{table}[h]
\centering
\caption{Per-class performance on the VisA dataset.}
\label{perclass_visa_results}
\resizebox{1\linewidth}{!}{
\begin{tabular}{@{}c|cc|cc@{}}
\toprule[1.5pt]
\rowcolor{gray!20} \multirow{2}{*}{Objects} & \multicolumn{2}{c|}{Pixel-level} & \multicolumn{2}{c}{Image-level} \\ 
\cmidrule(lr){2-3} \cmidrule(lr){4-5}
\rowcolor{gray!20} & AUROC (\%) & PRO (\%) & AUROC (\%) & AP (\%) \\
\midrule
Candle     & 99.09 & 95.62 & 88.79 & 91.41 \\
Capsules   & 97.61 & 90.87 & 80.53 & 89.44 \\
Cashew     & 92.69 & 93.61 & 90.20 & 95.54 \\
Chewinggum & 99.16 & 85.51 & 95.80 & 98.29 \\
Fryum      & 94.68 & 90.57 & 81.16 & 90.76 \\
Macaroni1  & 99.24 & 96.42 & 87.26 & 88.97 \\
Macaroni2  & 98.35 & 88.81 & 63.79 & 67.87 \\
PCB1       & 91.85 & 88.72 & 67.87 & 74.15 \\
PCB2       & 92.17 & 78.38 & 80.28 & 81.05 \\
PCB3       & 87.55 & 80.23 & 78.71 & 81.62 \\
PCB4       & 94.53 & 86.00 & 94.92 & 95.11 \\
Pipe Fryum & 96.20 & 96.32 & 90.48 & 95.36 \\
\midrule
Mean       & 95.26 & 89.26 & 83.32 & 87.46 \\
\bottomrule[1.5pt]
\end{tabular}
}
\end{table}

\noindent \textbf{Image-level AP.}~ Average Precision (AP) at the image level measures the model’s performance in ranking images based on their anomaly scores. It is computed as the area under the Precision-Recall (PR) curve, where precision and recall are defined as:
\begin{equation}
    Precision = \frac{TP}{TP + FP}, \quad Recall = \frac{TP}{TP + FN}
\end{equation}
\noindent Unlike AUROC, which considers both positive and negative classes, AP focuses on the positive (anomalous) class, making it particularly useful when anomalies are rare. A higher AP score indicates that the model ranks anomalous images higher than normal ones, ensuring effective anomaly prioritization in real-world applications.

\subsection{Objective Functions}

To optimize our anomaly detection model, we utilize Dice Loss~\cite{dice} and Focal Loss~
\cite{focal}. These loss functions are designed to handle class imbalance and improve segmentation accuracy.

\begin{table}
\centering
\caption{Per-class performance on the MPDD dataset.}
\label{perclass_mpdd_results}
\resizebox{1\linewidth}{!}{
\begin{tabular}{@{}c|cc|cc@{}}
\toprule[1.5pt]
\rowcolor{gray!20} \multirow{2}{*}{Objects} & \multicolumn{2}{c|}{Pixel-level} & \multicolumn{2}{c}{Image-level} \\ 
\cmidrule(lr){2-3} \cmidrule(lr){4-5}
\rowcolor{gray!20} & AUROC (\%) & PRO (\%) & AUROC (\%) & AP (\%) \\
\midrule
Bracket (Black) & 95.76 & 84.63 & 49.93 & 57.71 \\
Bracket (Brown) & 92.39 & 82.17 & 61.24 & 74.34 \\
Bracket (White) & 98.94 & 96.04 & 65.78 & 67.24 \\
Connector       & 96.71 & 89.15 & 91.19 & 88.22 \\
Metal Plate    & 94.52 & 88.00 & 77.84 & 91.38 \\
Tubes          & 98.82 & 95.72 & 96.47 & 98.47 \\
\midrule
Mean           & 96.19 & 89.28 & 73.74 & 79.56 \\
\bottomrule[1.5pt]
\end{tabular}
}
\end{table}

\begin{table}
\centering
\caption{Per-class performance on the BTAD dataset.}
\label{perclass_btad_results}
\resizebox{0.9\linewidth}{!}{
\begin{tabular}{@{}c|cc|cc@{}}
\toprule[1.5pt]
\rowcolor{gray!20} \multirow{2}{*}{Objects} & \multicolumn{2}{c|}{Pixel-level} & \multicolumn{2}{c}{Image-level} \\ 
\cmidrule(lr){2-3} \cmidrule(lr){4-5}
\rowcolor{gray!20} & AUROC (\%) & PRO (\%) & AUROC (\%) & AP (\%) \\
\midrule
01   & 93.97 & 76.63 & 94.17 & 97.96 \\
02   & 92.54 & 60.99 & 76.17 & 95.99 \\
03   & 94.39 & 89.20 & 99.62 & 96.69 \\
\midrule
Mean & 93.63 & 75.61 & 89.98 & 96.88 \\
\bottomrule[1.5pt]
\end{tabular}
}
\end{table}

\begin{table}
\centering
\caption{Per-class performance on the SDD dataset.}
\label{perclass_sdd_results}
\resizebox{0.8\linewidth}{!}{
\begin{tabular}{@{}c|cc|cc@{}}
\toprule[1.5pt]
\rowcolor{gray!20} \multirow{2}{*}{Objects} & \multicolumn{2}{c|}{Pixel-level} & \multicolumn{2}{c}{Image-level} \\ 
\cmidrule(lr){2-3} \cmidrule(lr){4-5}
\rowcolor{gray!20} & AUROC (\%) & PRO (\%) & AUROC (\%) & AP (\%) \\
\midrule
SDD   & 96.83 & 94.88 & 92.21 & 68.89 \\
\midrule
Mean  & 96.83 & 94.88 & 92.21 & 68.89 \\
\bottomrule[1.5pt]
\end{tabular}
}
\end{table}

\begin{table}
\centering
\caption{Per-class performance on the DTD-Synthetic dataset.}
\label{perclass_dtd_results}
\resizebox{1\linewidth}{!}{
\begin{tabular}{@{}c|cc|cc@{}}
\toprule[1.5pt]
\rowcolor{gray!20} \multirow{2}{*}{Objects} & \multicolumn{2}{c|}{Pixel-level} & \multicolumn{2}{c}{Image-level} \\ 
\cmidrule(lr){2-3} \cmidrule(lr){4-5}
\rowcolor{gray!20} & AUROC (\%) & PRO (\%) & AUROC (\%) & AP (\%) \\
\midrule
Blotchy\_099     & 99.45 & 95.94 & 66.56 & 90.76 \\
Fibrous\_183     & 99.44 & 97.89 & 86.56 & 96.60 \\
Marbled\_078     & 99.30 & 97.23 & 85.88 & 96.09 \\
Matted\_069      & 95.49 & 78.34 & 58.67 & 86.23 \\
Mesh\_114        & 95.48 & 82.55 & 87.12 & 94.52 \\
Perforated\_037  & 94.68 & 90.60 & 96.06 & 99.03 \\
Stratified\_154  & 98.92 & 98.62 & 96.25 & 99.17 \\
Woven\_001       & 99.82 & 99.04 & 99.38 & 99.76 \\
Woven\_068       & 99.02 & 96.94 & 97.32 & 98.58 \\
Woven\_104       & 98.79 & 96.69 & 99.31 & 99.85 \\
Woven\_125       & 99.76 & 98.90 & 100.00 & 100.00 \\
Woven\_127       & 94.39 & 90.15 & 92.55 & 94.41 \\
\midrule
Mean             & 97.88 & 93.58 & 88.81 & 96.25 \\
\bottomrule[1.5pt]
\end{tabular}
}
\end{table}

\begin{figure*}
\begin{center}
    \includegraphics[width=0.9\linewidth]{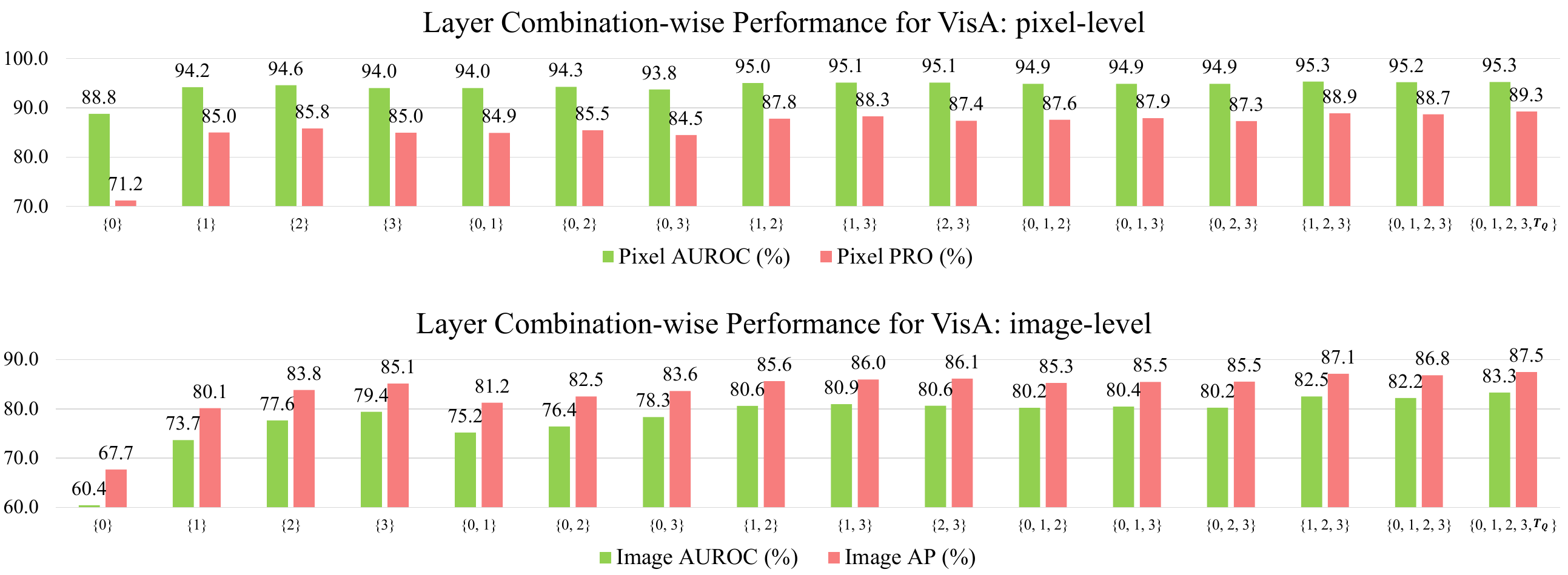}
\end{center}
   \caption{Performance comparison regarding different layer combinations, experimented with VisA dataset. We demonstrate the image-level performances for each combination. We demonstrate the image-level performances for each combination. $\{i\}$ indicates the $i_{th}$ layer block in the vision encoder, where $i \in \{1, 2, \cdots L\}$. All possible combinations are used for this experiment. }
\label{fig:layercombi_visa}
\end{figure*}

\noindent \textbf{Dice Loss} is derived from the Dice Similarity Coefficient (DSC), which measures the overlap between the predicted segmentation mask and the ground truth. It is particularly effective in cases where class imbalance exists, as it focuses on the relative overlap rather than absolute pixel counts. The Dice coefficient is defined as:

\begin{equation}
    DSC = \frac{2 |P \cap G|}{|P| + |G|},
\end{equation}
\noindent where \( P \) represents the set of predicted positive pixels, and \( G \) represents the set of ground truth positive pixels. The Dice Loss is formulated as:

\begin{equation}
    \mathcal{L}_{Dice} = 1 - \frac{2 \sum_{i} p_i g_i}{\sum_{i} p_i + \sum_{i} g_i + \epsilon},
\end{equation}
\noindent where \( p_i \) and \( g_i \) are the predicted and ground truth values for pixel \( i \), and \( \epsilon \) is a small constant to prevent division by zero. By optimizing this loss, the model learns to maximize the overlap between predictions and ground truth masks, which is crucial for accurate anomaly localization.

\noindent \textbf{Focal Loss} is an extension of the standard cross-entropy loss designed to address class imbalance by down-weighting easy examples and focusing on hard-to-classify samples. It introduces a modulating factor to emphasize difficult samples, making it particularly useful in anomaly detection where anomalies are rare. The Focal Loss is defined as:
\begin{equation}
    \mathcal{L}_{Focal} = - \sum_{i} \alpha (1 - p_i)^\gamma g_i \log p_i + (1 - g_i) \log (1 - p_i),
\end{equation}
\noindent where \( p_i \) is the predicted probability for pixel \( i \), and \( g_i \) is the ground truth label (1 for anomalous, 0 for normal). The parameter \( \alpha \) is a weighting factor to balance positive and negative samples, and \( \gamma \) is the focusing parameter that reduces the weight of easy-to-classify examples. When \( \gamma = 0 \), Focal Loss simplifies to the standard cross-entropy loss. As \( \gamma \) increases, the model focuses more on misclassified examples. This is beneficial in anomaly detection, where anomalies are often underrepresented in training data.

\subsection{More Implementation Details}~\label{appendix:implementation}
For training, we employ the Adam optimizer~\cite{kingma2014adam} with an initial learning rate of $4e^{-5}$. The model is trained for $15$ epochs. The weighting factor $\alpha$ used for the ensemble of $\mathbf{S_V}$ and $\mathbf{S_Q}$ is set to $0.8$, while the $\sigma$ value for Gaussian smoothing is set to $9$. Additionally, the values of $N_1$ and $N_2$ used in calculating image score are set to $500$ and $2500$, respectively. The number of prompt tokens modified for text prompt tuning is one. All experiments are conducted using PyTorch on a single Nvidia RTX A5000 GPU. 

\subsection{Compared Methods}
\textbf{WinCLIP~\cite{winclip}.}~ WinCLIP represents a pioneering approach in leveraging CLIP for ZSAD. At its core, WinCLIP employed a Compositional Prompt Ensemble (CPE) to address the challenge of adapting CLIP to anomaly detection tasks. Specifically, it utilized a diverse array of normal and abnormal text templates to construct rich linguistic representations that capture both typical and atypical object states. By combining object categories with descriptive templates, WinCLIP generated a wide range of prompts to effectively align visual features with textual embeddings, enabling it to identify anomalies without requiring task-specific fine-tuning. These prompts were designed to exploit CLIP’s pre-trained language-vision alignment, ensuring that the model can generalize across various anomaly detection scenarios. For anomaly segmentation, WinCLIP further refined its predictions by employing a window-based dense representation technique that extracts localized feature embeddings aligned with the generated prompts.

While WinCLIP effectively harnessed CLIP's general capabilities for anomaly detection, it was inherently limited by the fact that CLIP was not originally designed for this task. This limitation highlighted the need for future work to develop models or adaptations specifically tailored to anomaly detection while retaining the generalization power of vision-language pre-trained models like CLIP.

\noindent \textbf{AnomalyCLIP~\cite{anomalyclip}.}~ AnomalyCLIP introduced a novel approach that leverages object-agnostic prompt learning. Unlike traditional methods that relied on object-specific semantics, AnomalyCLIP focused on learning generalized normal and abnormal embeddings that are independent of object classes. This was achieved through the use of two simple yet effective text prompt templates: one representing normality and the other representing abnormality. These prompts replace class-specific terms with the generic term ``object," enabling the model to generalize across diverse domains without being constrained by specific class semantics. To further enhance its capabilities, AnomalyCLIP employs global and local context optimization during training, which aligns textual and visual embeddings more effectively. By combining these embeddings with a diagonally prominent attention mechanism (DPAM), AnomalyCLIP generates more precise anomaly segmentation maps, capturing fine-grained visual details.

Despite its advantages, AnomalyCLIP faced limitations inherent to its reliance on generic text prompts. These prompts, while effective for broad generalization, may struggle to capture nuanced, class-specific cues critical for certain domains. This highlighted a trade-off between generality and specificity, suggesting that future work could explore hybrid approaches to balance these aspects more effectively.

\noindent \textbf{AdaCLIP~\cite{adaclip}.}~ AdaCLIP proposed hybrid learnable prompts, combining static prompts and dynamic prompts generated from the final-layer token embedding of the vision encoder. This approach aimed to enhance CLIP's adaptability for ZSAD. The static prompts are shared across all images to adapt CLIP for ZSAD, while dynamic prompts are generated for each test image, providing CLIP with dynamic adaptation capabilities. 


Unlike AdaCLIP, which leverages both static and dynamic prompts for adaptation, our proposed GenCLIP framework adopts a different strategy to enhance generalization in ZSAD. Instead of relying on dynamic prompts derived from vision encoder embeddings, GenCLIP introduces specific general query prompt tokens (GQPs) during inference that does not incorporate vision features or class names. This design choice allows the model to better identify outliers where class information is not beneficial for anomaly detection. Furthermore, while AdaCLIP aims to improve adaptability through dynamic prompt generation, GenCLIP focuses on learning robust general prompts that can effectively handle diverse datasets without requiring image-specific adaptation. Our approach utilizes multi-layer prompting to enrich textual prompts with visual information in a structured manner, enhancing feature-level prompt augmentation for anomaly detection. Additionally, GenCLIP employs a dual-branch inference mechanism to leverage complementary information from both MVPs and GQPs, leading to improved generalization across different types of anomalies.

\subsection{Detailed ZSAD performance}~\label{appendix:performance}
\noindent \textbf{Per-class anomaly detection and segmentation results.} In Tables~\ref{perclass_mvtec_results}-\ref{perclass_sdd_results}, we present the per-class anomaly detection and segmentation performance for all six industrial datasets. Our method consistently achieves high AUROC and PRO scores across various object categories, demonstrating its effectiveness in both pixel-level anomaly localization and image-level anomaly detection. While performance varies depending on the dataset complexity and object texture, the overall results indicate strong generalization capabilities across diverse industrial scenarios.



\begin{figure*}
\begin{center}
    \includegraphics[width=0.9\linewidth]{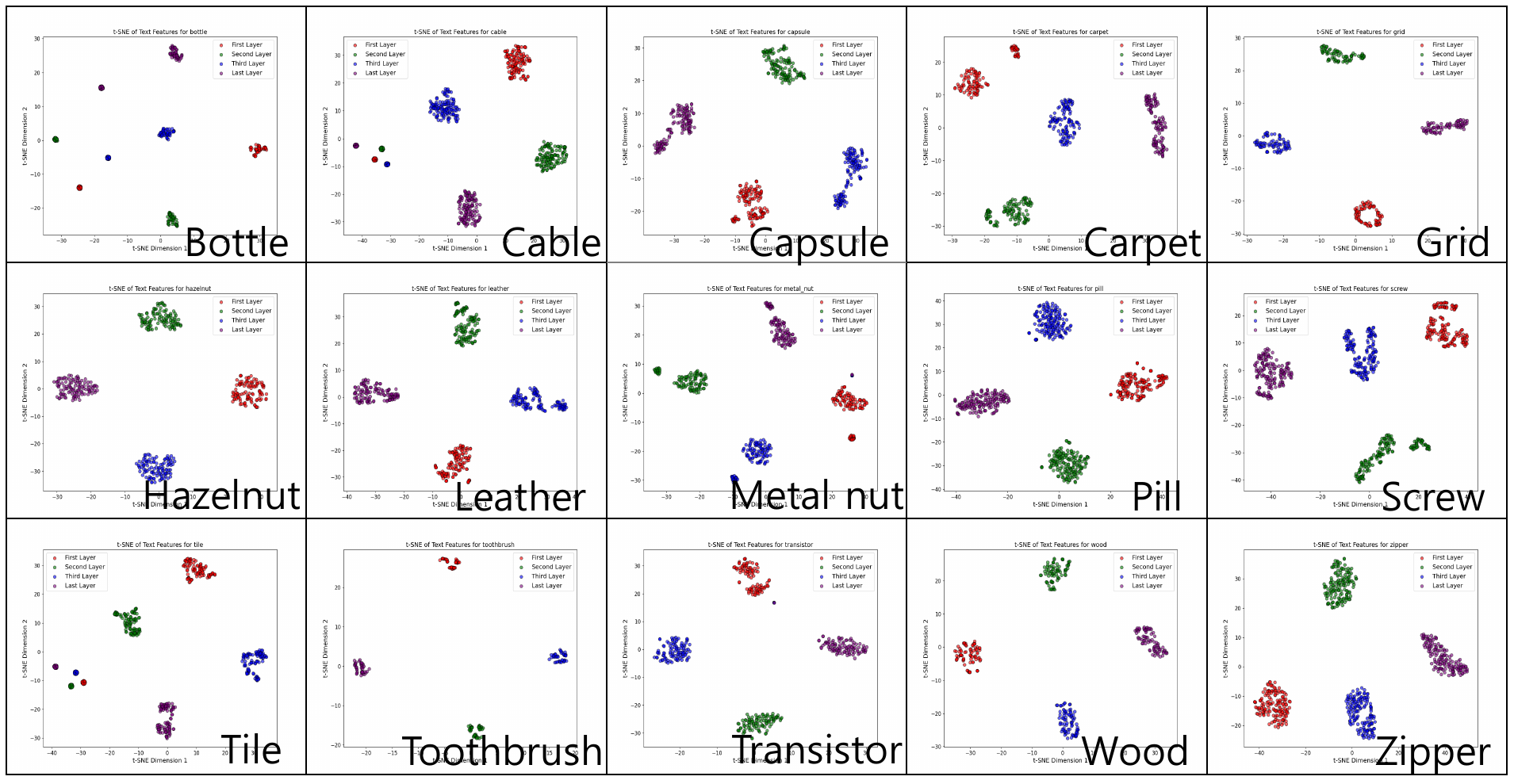}
\end{center}
   \caption{t-SNE visualization for MVTec object classes. Each layer of the text embeddings $\mathbf{F_T^i}$ is visualized. Red, green, blue, and purple indicate $\mathbf{F^0_T}$, $\mathbf{F^1_T}$ $\mathbf{F^2_T}$ $\mathbf{F^3_T}$. }
\label{fig:tsne_mvtec_layer}
\end{figure*}

\begin{figure*}
\begin{center}
    \includegraphics[width=0.9\linewidth]{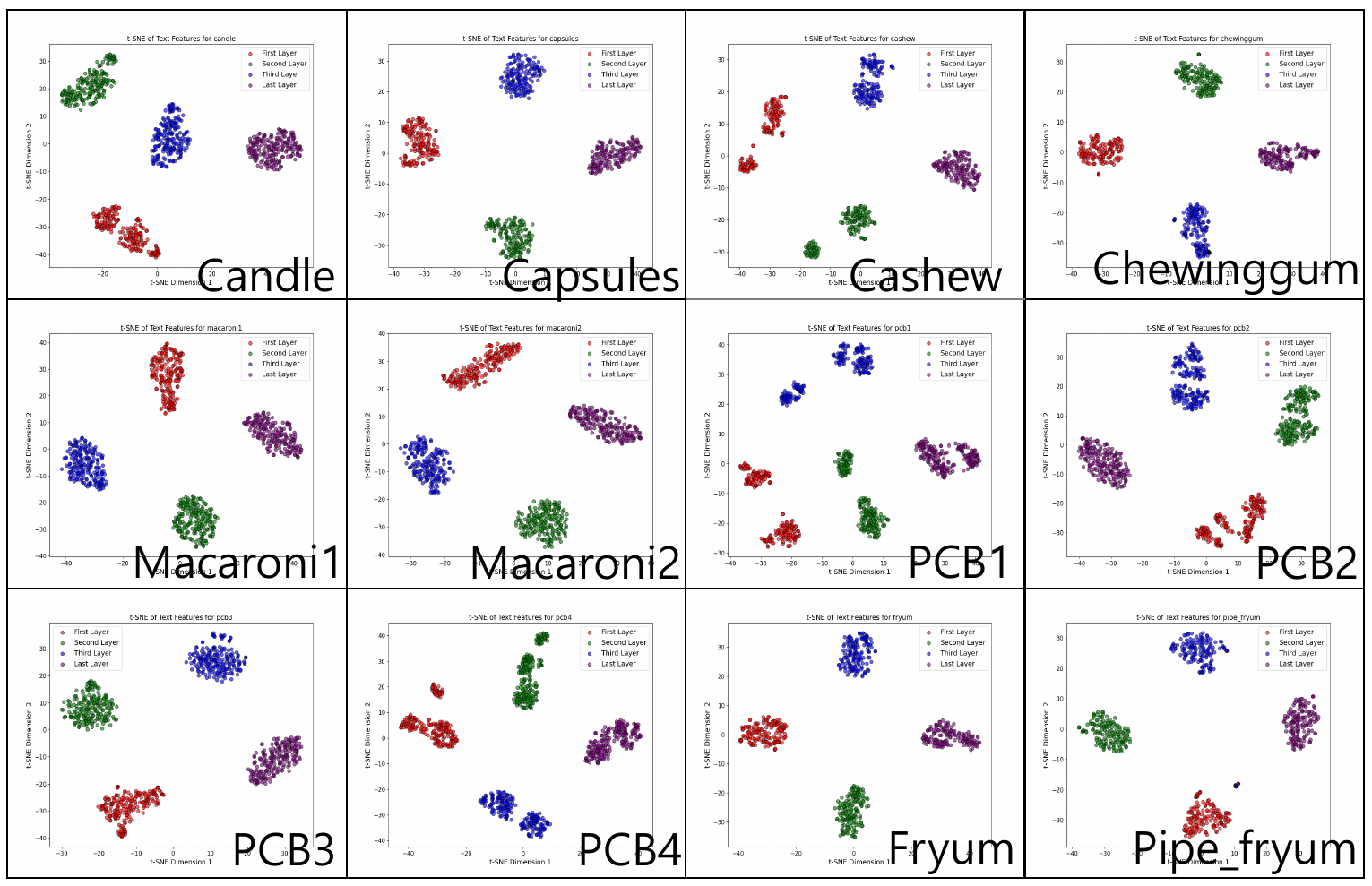}
\end{center}
   \caption{t-SNE visualization for VisA object classes. Each layer of the text embeddings $\mathbf{F_T^i}$ is visualized. Red, green, blue, and purple indicate $\mathbf{F^0_T}$, $\mathbf{F^1_T}$ $\mathbf{F^2_T}$ $\mathbf{F^3_T}$. }
\label{fig:tsne_visa_layer}
\end{figure*}

\begin{figure}[!h]
\begin{center}
    \includegraphics[width=0.8\linewidth]{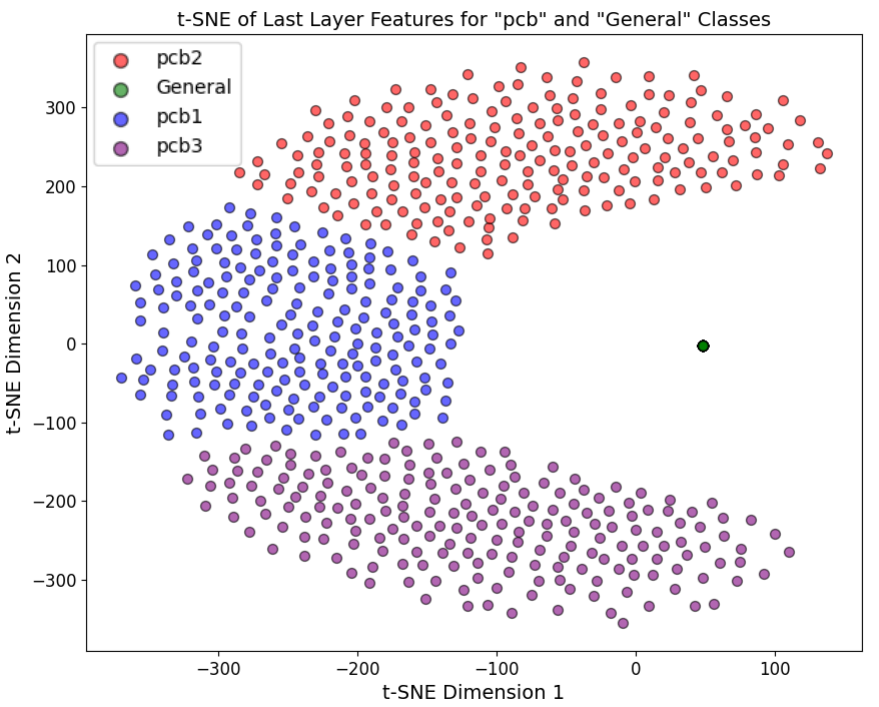}
\end{center}
   \caption{t-SNE visualization for specific classes of VisA that include ``pcb" in the label: ``pcb1", ``pcb2", ``pcb3". We also present the $\mathbf{F_Q}$ as well. Red, green, and purple are visualizations for text features of different PCB classes, and green indicates the GQP feature $\mathbf{F_Q}$. }
\label{fig:tsne_pcbs_general}
\end{figure}

\section{Analysis}

\subsection{Effects of MVP and GQP}~\label{appendix:layer}
In Fig.~\ref{fig:layercombi_visa} we present the performance evaluation of different layer combinations and $T_Q$ to validate the effectiveness of our layer-ensemble and two-branch inference method. Specifically, we assess four key metrics: pixel-level AUROC, PRO and image-level AUROC, AP for VisA. The results clearly demonstrate that leveraging multiple layers significantly enhances anomaly detection performance compared to using individual layers. Notably, the first layer exhibits the weakest capability in detecting anomalies, whereas middle layers contribute more effectively by capturing richer semantic features. Finally, we observe a noticeable performance improvement when adding a query-only branch that utilizes a highly general prompt, $T_Q$. This highlights the importance of integrating MVPs and GQPs to exploit diverse representations, ultimately improving robustness and accuracy across diverse datasets. 

Furthermore, Figs.~\ref{fig:tsne_mvtec_layer} and~\ref{fig:tsne_visa_layer} provide t-SNE visualizations of the text embeddings from different layers, generated from various classes in the MVTec and VisA datasets. In these visualizations, different colors represent different layers. The results confirm that each layer of GenCLIP's text embeddings captures distinct information, contributing to a more comprehensive representation of class-aware features for anomaly detection. This highlights the importance of leveraging MVPs to enhance the robustness and accuracy of anomaly detection models.

\begin{table}[]
\caption{Ablation studies regarding different generic terms for CNF, experimented with MVTec.}
\label{appendix:tab_cnfmvtec}
\centering
\resizebox{0.8\columnwidth}{!}{%
\begin{tabular}{c|cc|cc}
\hline
MVTec         & \multicolumn{2}{c|}{Pixel-level} & \multicolumn{2}{c}{Image-level} \\ \hline
Generic Term  & AUROC           & PRO          & AUROC  & \multicolumn{1}{c}{AP} \\ \hline
``object''      & \textcolor{red}{92.67}           & \textcolor{red}{88.11}          & 90.96  & 96.17                  \\
``manufacture" & \textcolor{red}{92.67}           & 88.08          & \textcolor{red}{90.98}  & \textcolor{red}{96.18}                  \\
``product"     & \textcolor{red}{92.67}           & 88.07          & \textcolor{red}{90.98}  & \textcolor{red}{96.18}                  \\
``item"        & \textcolor{red}{92.67}           & 88.07          & \textcolor{red}{90.98}  & \textcolor{red}{96.18}                  \\
``something"   & \textcolor{red}{92.67}           & 88.07          & \textcolor{red}{90.98}  & \textcolor{red}{96.18}                  \\ \hline
\end{tabular}}%
\end{table}

\begin{table}[]
\caption{Ablation studies regarding different generic terms for CNF, experimented with VisA.}
\label{appendix:tab_cnfvisa}
\centering
\resizebox{0.8\columnwidth}{!}{
\begin{tabular}{c|cc|cc}
\hline
VisA          & \multicolumn{2}{c|}{Pixel-level} & \multicolumn{2}{c}{Image-level} \\ \hline
General Term  & AUROC           & PRO          & AUROC  & \multicolumn{1}{c}{AP} \\ \hline
``object"      & 95.26           & \textcolor{red}{89.25}          & \textcolor{red}{83.31}  & \textcolor{red}{87.46}                  \\
``manufacture" & 95.26           & 89.10          & 82.90  & 87.24                  \\
``product"     & \textcolor{red}{95.30}           & 89.11          & 83.24  & 87.42                  \\
``item"        & 95.28           & 89.17          & 83.07  & 87.32                  \\
``something"   & \textcolor{red}{95.30}            & 89.13          & 83.15  & 87.37                  \\ \hline
\end{tabular}}
\end{table}
\subsection{Is the GQP Really Representative of Diverse Classes?}~\label{appendix:generaltsne}
Fig.~\ref{fig:tsne_pcbs_general} illustrates the effectiveness of our GQP in capturing a broad and representative feature space. Using t-SNE, we visualize the last-layer text embeddings of three PCB-related classes—-``pcb1," ``pcb2," and ``pcb3"—alongside $\mathbf{F_Q}$. As expected, the embeddings of the three PCB classes, extracted from their respective text features, form a tight cluster, reflecting their semantic similarity. Notably, the $\mathbf{F_Q}$ is positioned centrally within this cluster, indicating its ability to generalize across multiple PCB variations. This suggests that the $\mathbf{F_Q}$ effectively serves as an object-agnostic reference, capturing shared characteristics among related classes while maintaining adaptability for diverse downstream tasks. Such a property enhances its utility in ZSAD by providing a strong, generalized feature space that can accommodate variations across different object categories.

\subsection{Effects of Different Generic Terms for CNF}~\label{appendix:cnftext}
In Tabs.~\ref{appendix:tab_cnfmvtec}-\ref{appendix:tab_cnfsdd}, we report the performance of various experiments conducted with different generic terms for CNF. We selected five terms: ``object'', ``manufacture", ``product", ``item", and ``something". For most of the datasets, performance did not vary a lot. The dataset with the most noticeable variation was SDD, which contains texture-related images, rather than object images. In this case, the choice of general term had a greater impact compared to other datasets, where performance remained stable regardless of the term used. This suggests that in real-world industry applications, selecting terminology based on the characteristics of the classes can lead to more effective and contextually appropriate usage.
\begin{figure*}
\begin{center}
    \includegraphics[width=0.9\linewidth]{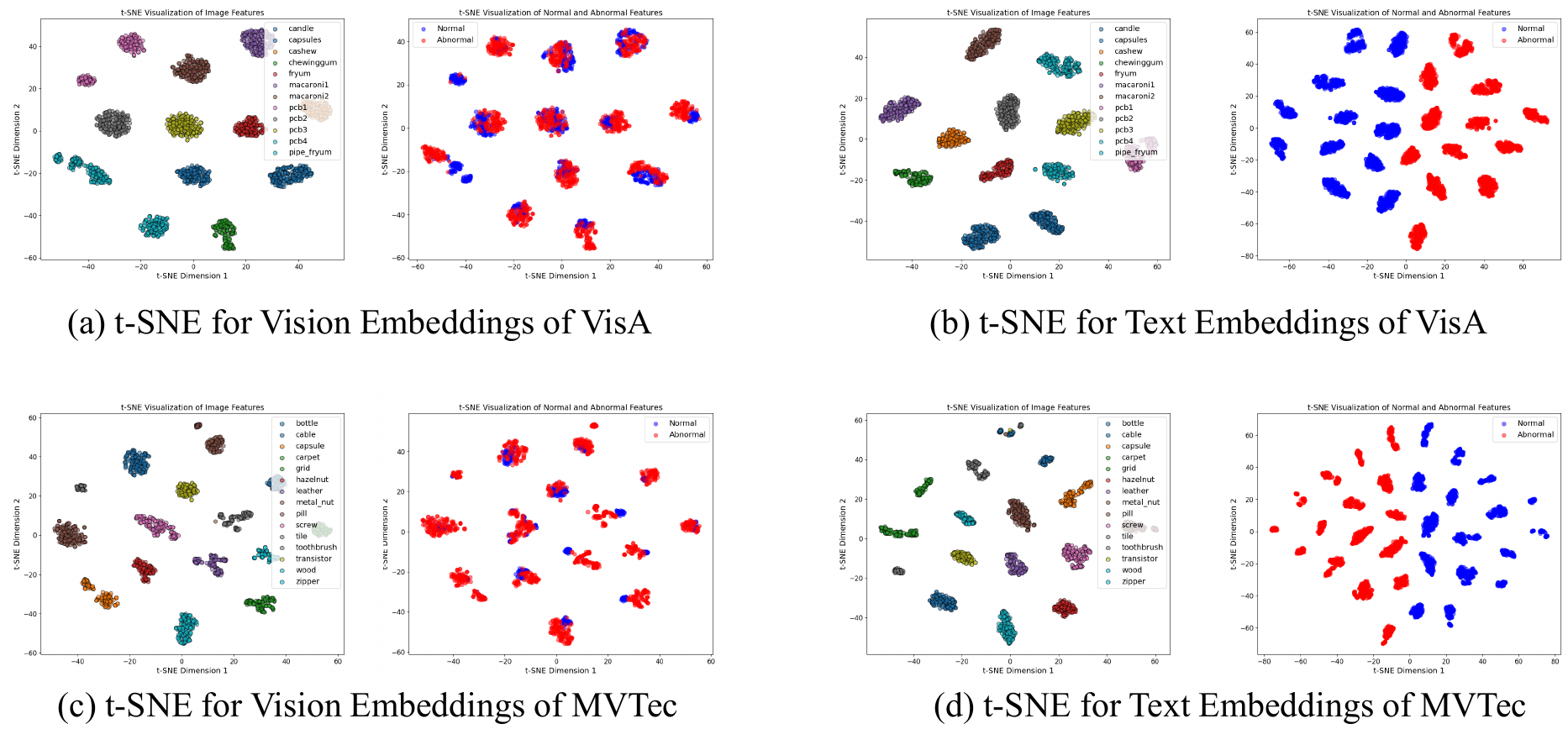}
\end{center}
   \caption{t-SNE visualization of image feature $\mathbf{F_I}$'s last layer for (a) MVTec and (b) VisA datasets. For each, the left images are organized object class-wise and the right images are organized by normal/abnormal labels. The blue and red indicate normal and abnormal, respectively.}
\label{fig:tsne}
\end{figure*}

\begin{table}[]
\label{appendix:tab_cnfbtad}
\caption{Ablation studies regarding different generic terms for CNF, experimented with BTAD.}
\centering
\resizebox{0.8\columnwidth}{!}{
\begin{tabular}{c|cc|cc}
\hline
BTAD           & \multicolumn{2}{c|}{Pixel-level} & \multicolumn{2}{c}{Image-level} \\ \hline
General Term   & AUROC           & PRO          & AUROC  & \multicolumn{1}{c}{AP} \\ \hline
``object"      & 93.63           & \textcolor{red}{75.61}          & 89.98  & \textcolor{red}{96.88}                  \\
``manufacture" & \textcolor{red}{93.65}           & 75.48          & 89.74  & 96.44                  \\
``product"     & 93.64           & 75.50          & 89.74  & 96.44                  \\
``item"        & 93.64           & 75.60          & \textcolor{red}{90.08}  & 96.51                  \\
``something"   & 93.00           & 73.19          & 88.97  & 96.81                  \\ \hline
\end{tabular}}
\end{table}

\begin{table}[]
\caption{Ablation studies regarding different generic terms for CNF, experimented with MPDD.}

\label{appendix:tab_cnfmpdd}
\centering
\resizebox{0.8\columnwidth}{!}{
\begin{tabular}{c|cc|cc}
\hline
MPDD           & \multicolumn{2}{c|}{Pixel-level} & \multicolumn{2}{c}{Image-level} \\ \hline
General Term   & AUROC           & PRO          & AUROC  & \multicolumn{1}{c}{AP} \\ \hline
``object"      & 96.19           & 89.23          & \textcolor{red}{73.86}  & \textcolor{red}{79.63}                  \\
``manufacture" & 96.18           & 89.19          & 73.85  & 79.47                  \\
``product"     & 96.17           & 89.16          & 73.85  & 79.47                  \\
``item"        & \textcolor{red}{96.20}            & \textcolor{red}{89.25}          & 73.31  & 78.89                  \\
``something"   & 96.16           & 89.15          & 73.85  & 79.47                  \\ \hline
\end{tabular}}
\end{table}

\begin{table}[]
\caption{Ablation studies regarding different generic terms for CNF, experimented with SDD.}
\label{appendix:tab_cnfsdd}
\centering
\resizebox{0.8\columnwidth}{!}{
\begin{tabular}{c|cc|cc}
\hline
SDD            & \multicolumn{2}{c|}{Pixel-level} & \multicolumn{2}{c}{Image-level} \\ \hline
General Term   & AUROC           & PRO          & AUROC  & \multicolumn{1}{c}{AP} \\ \hline
``object"      & 96.83           & 94.88          & 92.21  & 68.89                  \\
``manufacture" & 96.73           & 94.18          & \textcolor{red}{92.79}  & \textcolor{red}{72.71}                  \\
``product"     & 96.88           & \textcolor{red}{94.94}          & 92.34  & 72.65                  \\
``item"        & \textcolor{red}{96.98}           & 94.88          & 92.35  & 70.20                   \\
``something"   & 96.96           & 94.80           & 92.59  & 70.35                  \\ \hline
\end{tabular}}
\end{table}

\begin{table}[]
\label{appendix:tab_cnfdtd}
\caption{Ablation studies regarding different generic terms for CNF, experimented with DTD-Synthetic.}
\centering
\resizebox{0.8\columnwidth}{!}{
\begin{tabular}{c|cc|cc}
\hline
DTD            & \multicolumn{2}{c|}{Pixel-level} & \multicolumn{2}{c}{Image-level} \\ \hline
General Term   & AUROC           & PRO          & AUROC  & \multicolumn{1}{c}{AP} \\ \hline
``object"      & 97.87           & 93.57          & 88.83  & 96.25                  \\
``manufacture" & \textcolor{red}{97.90}            & 93.53          & 88.82  & 96.22                  \\
``product"     & 97.86           & 93.51          & 88.80   & 96.23                  \\
``item"        & 97.84           & 93.46          & \textcolor{red}{89.06}  & \textcolor{red}{96.31}                   \\
``something"   & 97.85           & \textcolor{red}{93.59}          & 88.34  & 96.11                  \\ \hline
\end{tabular}}
\end{table}

\subsection{t-SNE Visualization}

Fig.~\ref{fig:tsne} presents the feature visualization for the MVTec and VisA test datasets. Specifically, we extract features from the final layer of both our vision and text encoders to visualize vision and text embeddings, respectively. We then apply t-SNE for dimensionality reduction, projecting the features onto a 2D space. The results show that while the frozen vision encoder effectively captures object-specific semantic features, it struggles to model normality within images due to its original pretraining for object classification. In contrast, our text encoder integrates trainable tokens with vision tokens within the CLIP layer, which was initially pretrained for classification. This design enables our model to not only better capture normality but also maintain strong class discriminability by leveraging learnable tokens that enhance the alignment between textual and visual representations, ultimately improving anomaly detection. 

\subsection{Additional Qualitative Visualization}
\begin{figure*}
\begin{center}
    \includegraphics[width=0.9\linewidth]{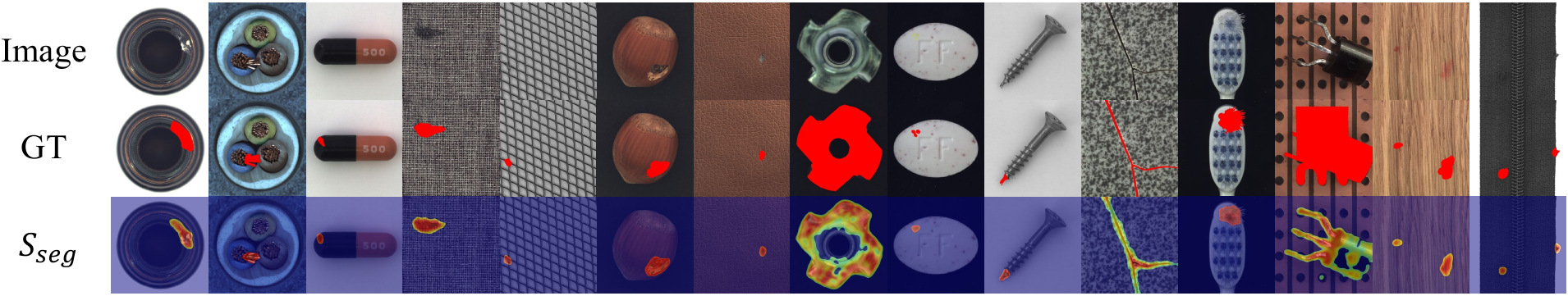}
\end{center}
   \caption{Visualization of GenCLIP's anomaly segmentation maps of abnormal images for MVTec dataset..}
\label{fig:vis_mvtec_abnormal}
\end{figure*}

\begin{figure*}
\begin{center}
    \includegraphics[width=0.9\linewidth]{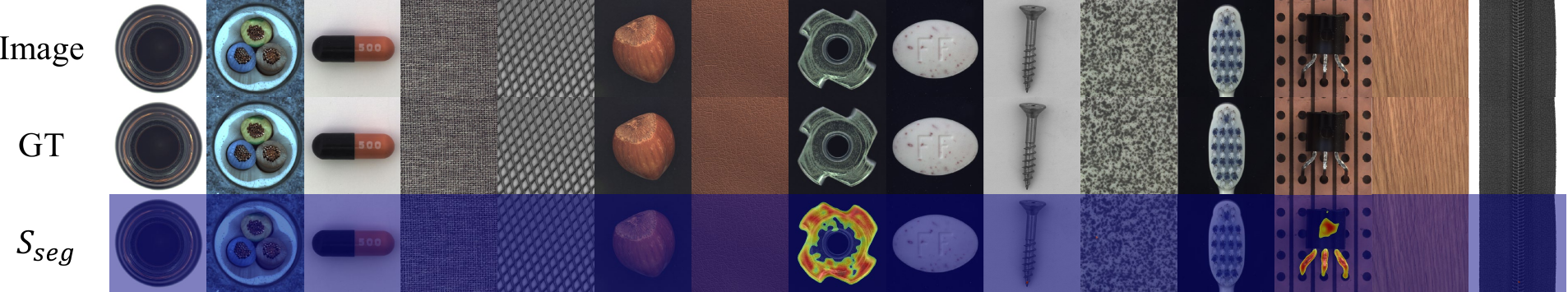}
\end{center}
   \caption{Visualization of GenCLIP's anomaly segmentation maps of normal images for MVTec dataset..}
\label{fig:vis_mvtec_normal}
\end{figure*}

\begin{figure*}
\begin{center}
    \includegraphics[width=0.9\linewidth]{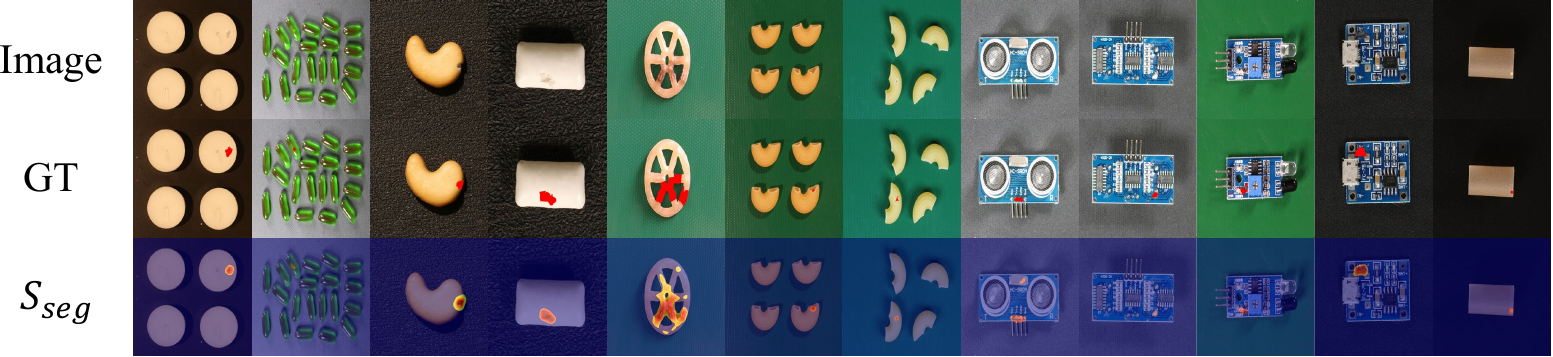}
\end{center}
   \caption{Visualization of GenCLIP's anomaly segmentation maps of abnormal images for VisA dataset.}
\label{fig:vis_visa_abnormal}
\end{figure*}

\begin{figure*}
\begin{center}
    \includegraphics[width=0.9\linewidth]{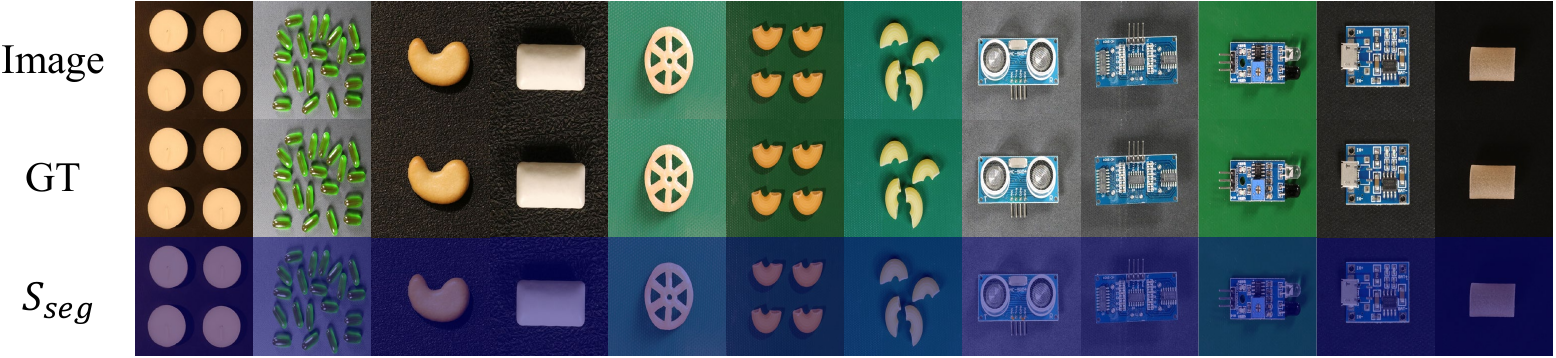}
\end{center}
   \caption{Visualization of GenCLIP's anomaly segmentation maps of normal images for VisA dataset.}
\label{fig:vis_visa_normal}
\end{figure*}

\begin{figure*}
\begin{center}
    \includegraphics[width=0.9\linewidth]{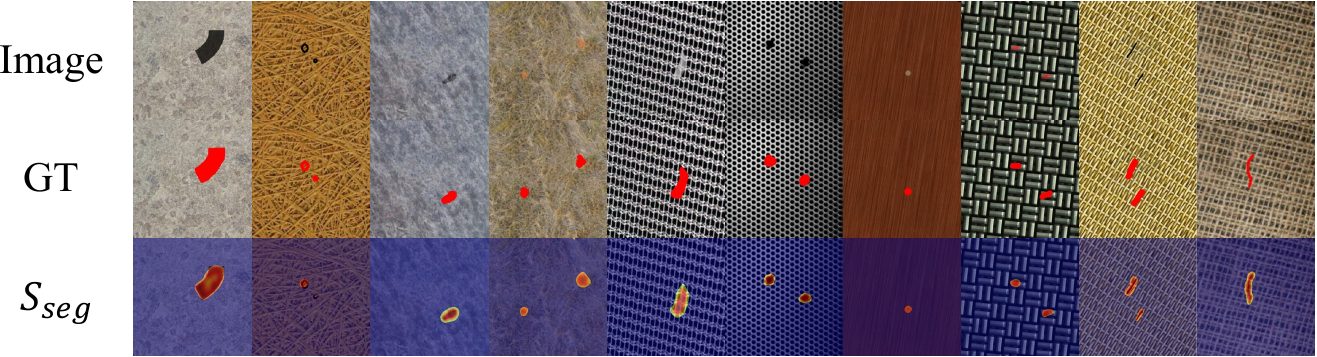}
\end{center}
   \caption{Visualization of GenCLIP's anomaly segmentation maps of abnormal images for BTAD dataset.}
\label{fig:vis_btech_abnormal}
\end{figure*}

\begin{figure*}
\begin{center}
    \includegraphics[width=0.9\linewidth]{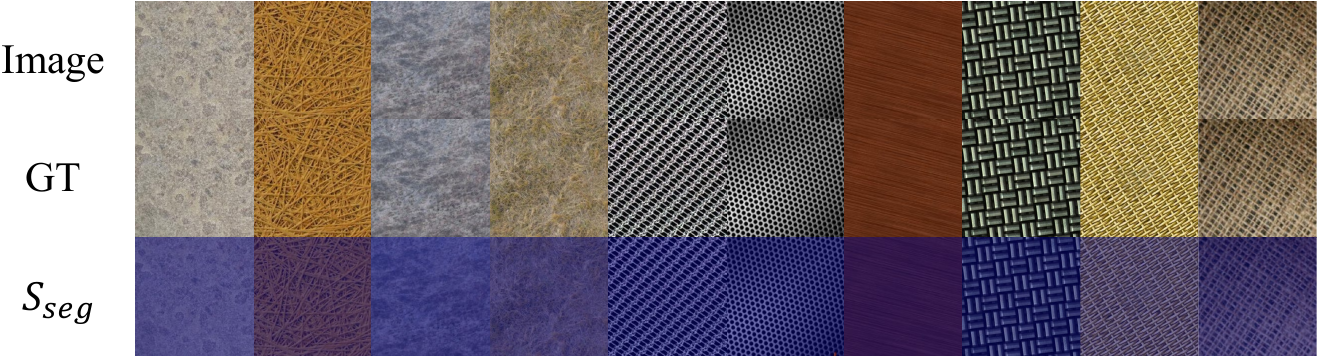}
\end{center}
   \caption{Visualization of GenCLIP's anomaly segmentation maps of normal images for BTAD dataset.}
\label{fig:vis_btech_normal}
\end{figure*}

\begin{figure*}
\begin{center}
    \includegraphics[width=0.9\linewidth]{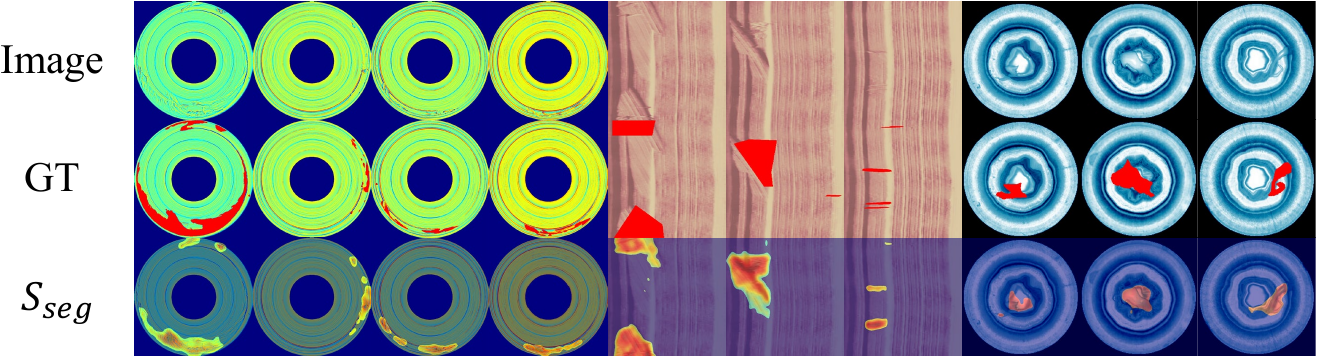}
\end{center}
   \caption{Visualization of GenCLIP's anomaly segmentation maps of abnormal images for DTD-Synthetic dataset.}
\label{fig:vis_dtd_abnormal}
\end{figure*}

\begin{figure*}
\begin{center}
    \includegraphics[width=0.9\linewidth]{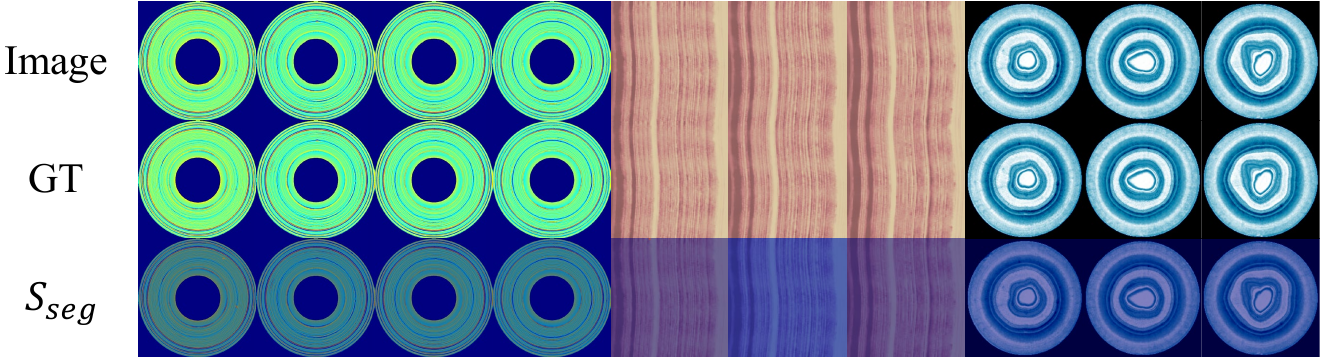}
\end{center}
   \caption{Visualization of GenCLIP's anomaly segmentation maps of normal images for DTD-Synthetic dataset.}
\label{fig:vis_dtd_normal}
\end{figure*}

\begin{figure*}
\begin{center}
    \includegraphics[width=0.9\linewidth]{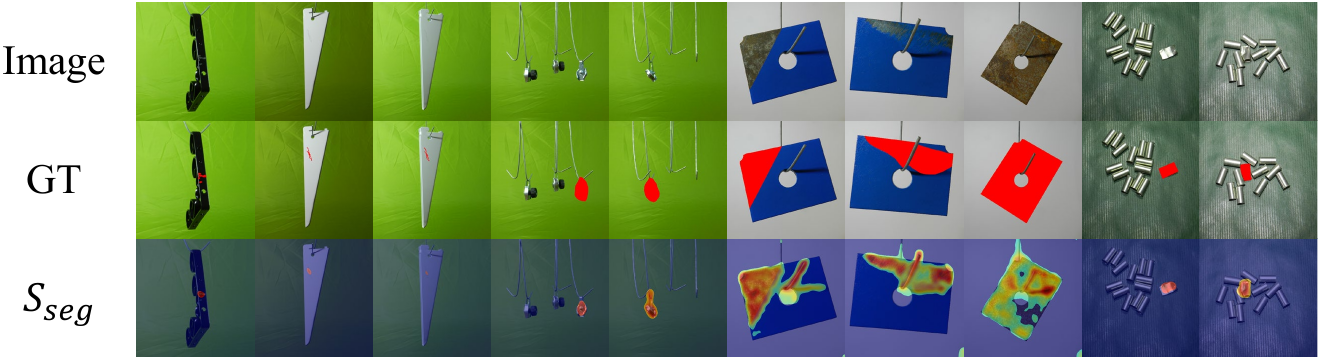}
\end{center}
   \caption{Visualization of GenCLIP's anomaly segmentation maps of abnormal images for MPDD dataset.}
\label{fig:vis_mpdd_abnormal}
\end{figure*}

\begin{figure*}
\begin{center}
    \includegraphics[width=0.9\linewidth]{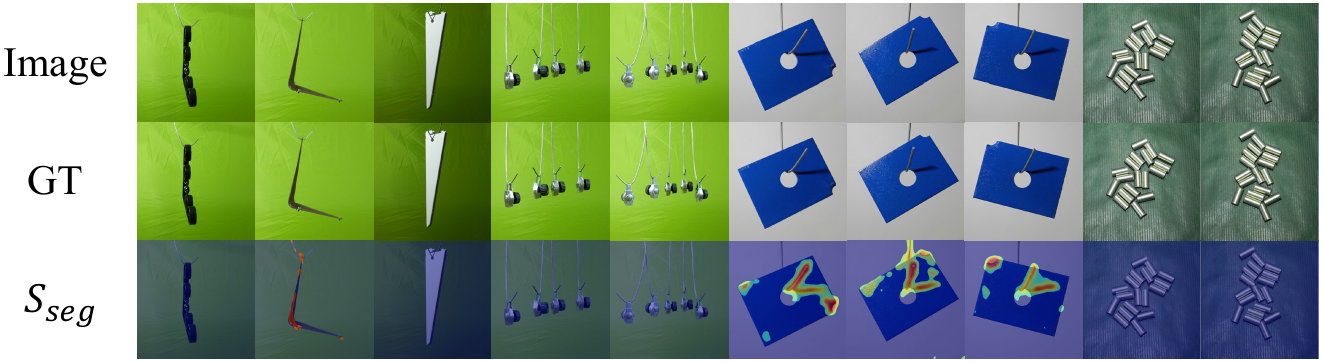}
\end{center}
   \caption{Visualization of GenCLIP's anomaly segmentation maps of normal images for MPDD dataset.}
\label{fig:vis_mpdd_normal}
\end{figure*}

\begin{figure*}
\begin{center}
    \includegraphics[width=0.9\linewidth]{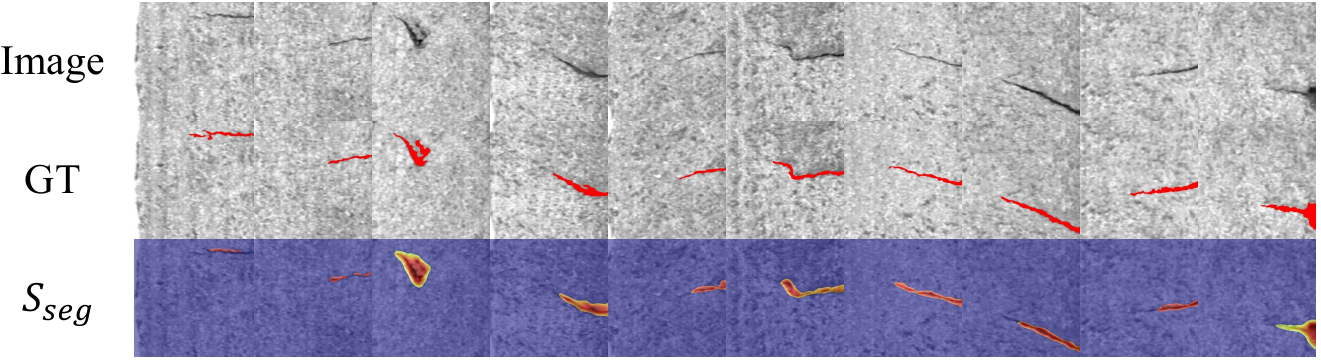}
\end{center}
   \caption{Visualization of GenCLIP's anomaly segmentation maps of abnormal images for SDD dataset.}
\label{fig:vis_sdd_abnormal}
\end{figure*}

\begin{figure*}
\begin{center}
    \includegraphics[width=0.9\linewidth]{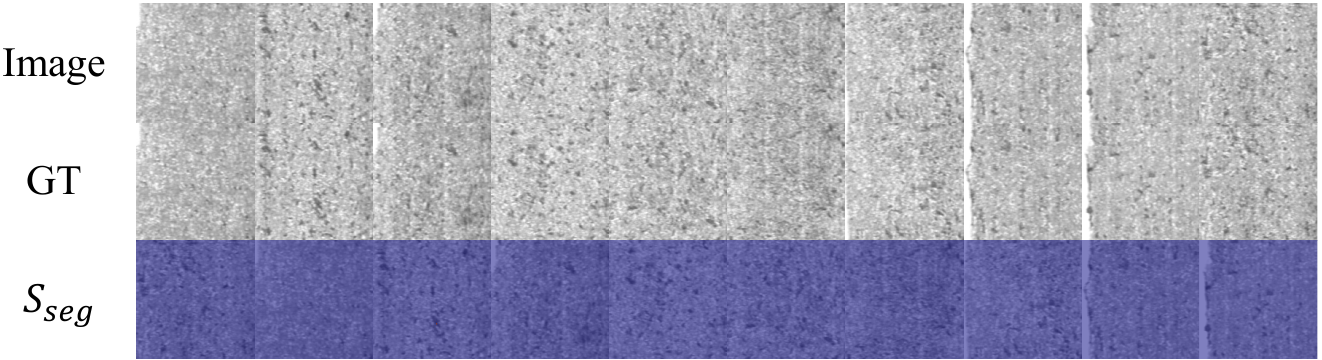}
\end{center}
   \caption{Visualization of GenCLIP's anomaly segmentation maps of normal images for SDD dataset.}
\label{fig:vis_sdd_normal}
\end{figure*}

In Figs.~\ref{fig:vis_mvtec_abnormal}-\ref{fig:vis_sdd_normal}, we present additional qualitative results across all six datasets used for inference. These figures showcase anomaly detection results for every object class in each dataset, demonstrating the versatility of our GenCLIP model. It can be observed that GenCLIP effectively detects anomalies across a wide range of industrial objects while also capturing diverse types of anomalies, including both structural defects and texture-based irregularities.

Furthermore, the results highlight the model’s ability to generalize across datasets with different anomaly characteristics, suggesting its robustness in real-world applications. The effectiveness of our approach is particularly evident in challenging cases where subtle anomalies need to be distinguished from normal variations. This adaptability underscores the potential of GenCLIP for practical deployment in anomaly detection tasks across various domains, ensuring comprehensive coverage of object classes within each dataset.

\end{document}